\documentclass[journal,twoside]{IEEEtran}

\usepackage{newtxtext}
\usepackage[varg]{newtxmath}

\usepackage{graphicx,xcolor}
\usepackage{amsfonts}
\usepackage{enumerate}
\newtheorem{thm}{Theorem}
\newtheorem{lem}{Lemma}

\newtheorem{pf}{Proof of Lemma}
\newtheorem{pft}{Proof of Theorem}

\ifCLASSINFOpdf
\else
\fi
\usepackage{url}

\begin{document}
%
\title{Approximate Spectral Decomposition of Fisher Information Matrix for Simple ReLU Networks}
%
%
%

\author{Yoshinari~Takeishi,~
        Masazumi~Iida,
        and~Jun'ichi~Takeuchi
\thanks{
}
\thanks{Y.\ Takeishi and J.\ Takeuchi 
are with the Faculty of Information Science and Electrical Engineering, 
 Kyushu University, Motooka 744, Nishi-ku, Fukuoka-city, 819--0395, Japan 
(e-mail: \{takeishi, tak\}@inf.kyushu-u.ac.jp).}
\thanks{M. Iida is with the Graduate School of Information Science and Electrical Engineering, 
 Kyushu University, Motooka 744, Nishi-ku, Fukuoka-city, 
  819--0395 Japan 
(e-mail: iida@me.inf.kyushu-u.ac.jp).}%

}

\maketitle

\begin{abstract}
We argue the Fisher information matrix (FIM) of one hidden layer networks with the ReLU activation 
function.
For a network,
let $W$ denote the $d \times p$ weight matrix from the 
$d$-dimensional input to the hidden layer consisting of $p$ neurons,
and $v$ the $p$-dimensional weight vector from the hidden layer to the scalar output.
We focus on the FIM of $v$, which we denote as $I$.
Under certain conditions, 
we characterize the first three clusters of eigenvalues and eigenvectors of the FIM.
Specifically, we show that 
1) Since $I$ is non-negative owing to the ReLU, the first eigenvalue
is the Perron-Frobenius eigenvalue.
2) For the cluster of the next maximum values, the eigenspace is spanned
by the row vectors of $W$.
3) 
The direct sum of the eigenspace of the first eigenvalue and that of the third cluster is spanned by the set of all the vectors obtained as the Hadamard product of any pair of the row vectors of $W$.
We confirmed by numerical calculation that the above is approximately correct
when the number of hidden nodes is about 10000.
\end{abstract}


%

\section{Introduction}
%
%
%
%


\IEEEPARstart{D}{eep}
neural networks show high generalization performance and have been very successful in recent years.
However, it is still not entirely clear why it works well.
The goal in learning network parameters is to reduce the generalization error.
The direction in the parameter space to reduce it can be found by examining the eigenvectors of 
the large eigenvalues of
its 
Fisher information matrix (FIM).
In order to investigate the behavior of generalization error, it is important to examine the FIM.

\begin{figure}[t]
\centering
\includegraphics[width=8.5cm]{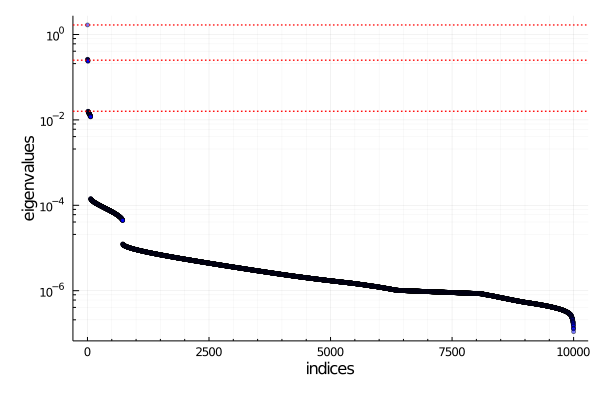}
\caption{Eigenvalues of $J^{(n)}$ for $n=100000$, $p=10000$, and $d=10$, which is proportional to the empirical FIM $I^{(n)}$ 
as described in Section \ref{FIMsec}}
\label{fig:d5_e}
\end{figure}

\begin{figure}[t]
\centering
\includegraphics[width=8.5cm]{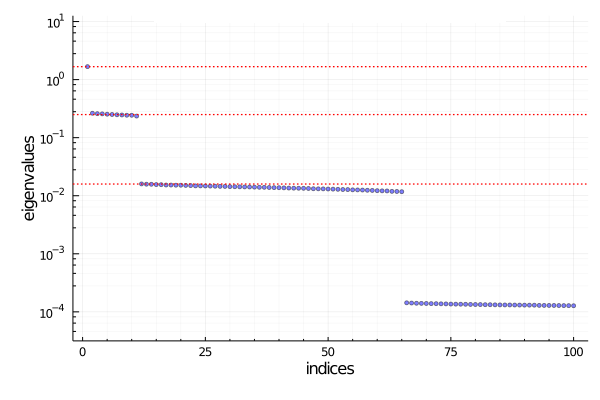}
\caption{The enlarged graph of the first 100 eigenvalues in Fig.~\ref{fig:d5_e}}
\label{fig:d5_e2}
\end{figure}

In this paper, we investigate a one hidden layer network with a ReLU activation function and 
consider only the last layer connection weight as a network parameter. We fix the weights other than the last layer.
In this setting, we calculated the eigenvalue distribution of the empirical FIM under certain conditions by numerical calculation, 
and obtained the result as shown in Fig.~\ref{fig:d5_e},
from which, we can see a remarkable ``grouping'' property regarding the magnitude of eigenvalues.
This phenomenon is related to the grouping of the eigenvalues of the Neural Tangent Kernel under certain conditions. The detail is discussed in Section \ref{Related_ntk}.
This paper 
investigates the first three clusters of eigenvalues and eigenvectors of the FIM.
Specifically, we show that 
the following approximately holds.
1) Since $I$ is non-negative owing to the ReLU, the first eigenvalue
is the Perron-Frobenius eigenvalue.
2) For the cluster of the next maximum values, the eigenspace is spanned
by the row vectors of $W$.
3) The direct sum of the eigenspace of the first eigenvalue and that of the third cluster is spanned by the set of all the vectors obtained as the Hadamard product of any pair of the row vectors of $W$.
Fig.~\ref{fig:d5_e2} is the enlarged graph of the first 100 eigenvalues in Fig.~\ref{fig:d5_e}.
We can confirm that it has the one remarkably large eigenvalue, the next largest group of $d$ eigenvalues, and 
the next largest group of $d(d+1)/2-1$ eigenvalues, where $d$ is the dimension of the input to the network.

This result is useful to understand the behavior of the gradient descent
in training the simple ReLU network.
In fact, the discussion in Section \ref{discuss} shows 
that the ``effective number of parameters'' grows as $O(1)$, $O(d)$, $O(d^2)$ when
we raise the goal for accuracy of estimation.
We further discuss in Section \ref{Gen_err} that in our case these eigenvectors are also in the direction of reducing the generalization error.

The grouping property comes from the fact that the ReLU activation function loses its linearity at the point where the input value is 0.
In investigating this phenomena, we found that Tian \cite{Tian} studied the same network as in our situation.
Using the result of his study, an explicit expression of the FIM was obtained.
It is not the form we wanted, but we succeeded to transform it to a form by which we can understand the situation of 
eigenvalue distribution we discovered. 

In order to match the realistic situation of deep learning, we need to consider multiple hidden layers networks (deep neural network; DNN)
and examine the FIM with the weights of all layers added as parameters.
Though we cannot analyze such DNNs in this paper, we think that our result reveals certain important properties of DNNs. 
Actually, Amari et al. \cite{Amari2} shows
that the FIM of DNN is approximately block diagonal, and 
the FIM we analyze can be regarded as a block of the FIM of DNN.

We explain the detailed structure of the neural network we analyze in Section \ref{sec2}.
The main result, the analysis of eigenvectors and eigenvalues of the FIM, is described in Section \ref{mainres} with some lemmas and theorems.
The results of numerical calculations are stated in Section \ref{simulate}.
Section \ref{discuss} discusses the behavior of the gradient descent by using our results on the FIM.

\section{Related works}

In this section, we discuss some related works.

\subsection{Related works on Neural Tangent Kernel}
\label{Related_ntk}
This study is closely related to the Neural Tangent Kernel (NTK) \cite{Jacot},
which is useful for investigating the learning behavior of DNNs 
when the width is infinitely large.
For shallow networks, such as those treated in this paper, the NTK can be computed in a simple form when the weights are initialized with an appropriate distribution.
In particular, letting the weight of the final layer be $v\in \Re^p$, and $x,x'\in \Re^d$ be inputs of the network, 
the NTK for $v$ in the limit of infinite width is given by \cite{Chizat}:
\begin{align*}
k^{(v)}(x,x')&=E_a[\varphi(a\cdot x)\varphi(a\cdot x')]\\
&=\frac{\|x\|\|x'\|E_a[\|a\|^2]}{2\pi d}((\pi-\theta)\cos\theta +\sin\theta),
\end{align*}
where the distribution of $a$ is rotation invariant in $\Re^d$, 
and $\theta\in[0,\pi]$ is the angle between $x$ and $x'$.
If we replace $x$ by the $i$-th column vector of $W$, $x'$ by the $j$-th column vector of $W$, and $a$ by the input $x$ in the above equation, 
we have exactly the $i,j$ components of our FIM. (See Lemma \ref{thm_fim} in this paper.)
We note that in this paper, each element of $x\in \Re^d$ is independently subject to the standard normal, which is a rotation invariant distribution.
Therefore, the generalization of the distribution of $x$ to a 
rotation invariant distribution would be straightforward.

Furthermore, if $x$ and $x'$ lie on the sphere $\mathbb{S}^{d-1}=\{x\in \Re^d : \|x\|=1\}$
, Bietti and Mairal \cite{Bietti} showed a decomposition of the 
NTK using a basis of spherical harmonics, and its eigenvalues have a grouping property.
Specifically, 
it has eigenvalues $\mu_0,\mu_1,\mu_2,\ldots$ and the dimension of the eigenspace of $\mu_k$ is 
$N(d,k)=\frac{2k+d-2}{k}\binom{k+d-3}{d-2}$ 
, where $\mu_k=0$ 
if $k$ is an odd number greater than or equal to 3.
This implies that if the column vectors of $W$ follow a uniform distribution over ${S}^{d-1}$, the eigenvalues of our FIM have the grouping property as $p$ grows large.
In fact, $N(d,1)=d$ and $N(d,2)=(d-1)(d+2)/2$ coincide with the number of eigenvalues in the second and third clusters in our FIM, respectively.
Note that in our analysis, the column vectors of $W$ do not follow a uniform distribution over ${S}^{d-1}$, but 
each element of the column vectors is independently generated from a Gaussian distribution with mean 0 and variance $1/p$
\footnote{
In \cite{Amari}, the variance is set to $\sigma_W^2/p$, and we adopt the case $\sigma_W=1$ in this paper. 
If one sets the variance to $\sigma_W^2/p$, the eigenvalues are $\sigma_W^2$ times larger.
On the other hand, in \cite{Lecun} and \cite{Jacot}, the variance is set to $1/d$, which also seems to be common.
}.
This difference would be small in high dimensions.
Therefore, we believe the novelty of our study lies 
in the characterization of the first three groups of the eigenvalues and eigenvectors, rather than the grouping of the eigenvalues.
It would also be noted that our result is shown assuming $p$ is finite,
while the analysis in \cite{Bietti} assumes $p$ is infinitely large.

\subsection{Other related works}
\label{Related}
Since 2010's it has been an important issue 
to find the reason why the overfitting problem does not occur 
for DNNs despite of their huge number of parameters.
By considering the NTK, Jacot et al. \cite{Jacot} 
showed that there exists an optimal solution near randomly chosen initial weight parameters, 
and Amari \cite{Amari} gave an intuitive interpretation of this result by high-dimensional geometry.
In Section 3 in \cite{Amari}, they investigated a one hidden layer network, which is the same one as our case.
The analysis for one hidden networks
is for the simplest case, but it contains an important aspect of the learning process of DNNs.

On the other hand, Belkin et al. \cite{Belkin} have addressed this overfitting problem from the phenomenon 
called ``double descent'' of the generalization error.
In particular, they showed in \cite{Belkin2} that this phenomenon occurs in a certain linear regression problem setting 
using the least squares estimator.
As shown in Section \ref{sec2}, our case can also be considered as a linear regression problem, 
and is similar to their case where double descent occurs.

Another approach is to investigate the learning of DNNs from the perspective of 
the minimum description length (MDL) principle \cite{Ress}.
In MDL principle, a model that can express given data with the shortest description length should be selected.
Barron and Cover \cite{Barron} showed that in unsupervised learning, selecting a model according to the MDL principle 
enjoys a small generalization error.
Recently, Kawakita and Takeuchi \cite{Kawakita} have shown that this result can be extended to supervised learning, 
and it might also be extended to deep neural networks.
Although they assumed that the design matrix of linear regression follows a Gaussian distribution, 
our work might provide an understanding of the case where the design matrix is restricted in a lower dimensional space determined 
by the activation function.

Finally, we introduce a few studies that investigate the eigenvalues of the FIM of neural networks.
LeCun et al. \cite{Lecun} empirically reported the existence of a few very large eigenvalues 
compared to the others for the Hessian of the loss function.
In recent years, Karakida et al. \cite{Karakida} have investigated the mean, variance, and maximum of eigenvalues
of FIM at infinite width using mean field theory.
In our study, we investigate not only the maximum eigenvalues, 
but also the main eigenvalues and their eigenvectors in more limited cases.

\section{Preliminaries}
\label{sec2}
In this section, we describe the linear regression problem with a one hidden layer network
 and introduce the Fisher information matrix (FIM), which is the main research subject of this paper.
Although the linear regression problem itself and the empirical FIM are not the main research topics of this paper, we introduce them here in order to discuss the behavior of gradient descent in Section \ref{discuss}.
We also note that the linear regression problem we discuss is the same as that discussed in Section 3 in \cite{Amari}, for instance.  


\subsection{One hidden layer network}
\label{onehid}
We consider a one hidden layer network, which has $p$ neurons in its hidden layer.
Let $x=(x_1,\ldots,x_d)\in \Re^d$ be an input to the network.
We assume that $p >> d$ and $d$ is also sufficiently large.
The output of the $j$-th neuron $X_j\in \Re$ is written as 
\begin{eqnarray*}
X_j= \varphi\left(\sum_{i=1}^{d} x_i W_{ij}\right) \ \text{for $j$ = 1,2,\dots,$p$},
\end{eqnarray*}
where $W \in \Re^{d\times p} $ is a weight matrix, and the activation function $\varphi$ is defined as
\begin{eqnarray*}
\varphi(z)= 
\begin{cases}
        z   & \ \text{if $z \geq 0$},\\
        0   & \ \text{if $z < 0$}
\end{cases}
\end{eqnarray*}
for $z\in \Re$.
In this paper, we consider a setting without the bias term in the network.
The output of the network $y\in \Re$ is 
\begin{eqnarray*}
y=Xv^{*T} + \epsilon,
\end{eqnarray*}
where  $X$ is the vector $(X_1,\ldots,X_p)$, 
$v^* \in \Re^p $ a connection weight vector ($p$-dimensional row vector), 
and $\epsilon\in\Re$ a noise subject to $N(0, \sigma^2)$.
Here $N(\mu,\sigma^2)$ denotes the Gaussian distribution with mean $\mu$ and variance $\sigma^2$.

Let $W$ be fixed and consider the problem of estimating $v^*$ from $n$ training data
$\left\{x_{(1)},y_{(1)}\right\}, \ldots,\left\{x_{(n)},y_{(n)}\right\}$,
where each $x_{(t)}$ is independently subject to the $d$-dimensional standard normal distribution.
This can be regarded as a linear regression problem with $X$ as the explanatory variable, 
where $X$ is distributed in the $d$-dimensional subspace
in $\Re^p$
determined by the matrix $W$ and 
the ReLU activation function.

In this paper, we make an additional assumption that all columns of matrix $W$ are non-zero vectors. This is natural because when the $j$-th column of $W$ is a zero vector, the $j$-th neuron will have no contribution to the network's output for any input $x$. 

\subsection{Generalization error}
\label{Gen_err}
Hereafter, $v$ denotes the estimate of $v^*$.
In the above setting, we want to find $v$ with a small generalization error.
Let $J$ denote the $p\times p$ matrix $E[X^TX]$, where 
the expectation is taken for $x$.
Recall that $x$ is independently subject to the $d$-dimensional standard normal distribution as stated in previous subsection
.
Then, the generalization error is
\begin{eqnarray*}
E[(y-Xv^T)^2] &=& E[(X(v^*-v)^T+\epsilon)^2] \\
 &=& (v-v^{*})J(v-v^{*})^T+\sigma^2,
\end{eqnarray*}
where the expectation is taken for $x$ and $\epsilon$.
Considering eigenvectors of $J$, 
we understand that the generalization error can be reduced in the direction of 
eigenvectors which correspond large eigenvalues. 
Therefore, 
it is important to find these eigenvectors and eigenvalues.
The next section describes our findings about it.

{\it Remark}: If there is no activation function, that is, $X= xW$, the situation is simple.
Namely, we have only $d$ eigenvectors of which directions match $W_1, W_2, \ldots, W_d$, 
where $W_l$ denotes the $l$-th row vector of $W$.
Non-linearity of ReLU complicates this problem.

\subsection{Fisher information matrix}
\label{FIMsec}
Let $p(x,y|v)$ denote the joint probability density of $(x,y)$ for $y=Xv^T + \epsilon$.
The Fisher information matrix (FIM) $I$ for $v$ is 
\begin{align*}
I_{ij} &=E\left[\frac{\partial}{\partial v_i}\log p(x,y|v) \frac{\partial}{\partial v_j}\log p(x,y|v) \right]\\
&= -E\left[\frac{\partial^2}{\partial v_i\partial v_j}\log p(x,y|v) \right].
\end{align*}
Since
\begin{align*}
\frac{\partial}{\partial v_i}
\log p(y|x,v) = 
\frac{\partial}{\partial v_i} \Bigl( -\frac{(y-Xv^T)^2}{2\sigma^2}\Bigr)
=\frac{(y-Xv)X_i}{\sigma^2},
\end{align*}
and
\begin{align*}
\frac{\partial^2}{\partial v_j \partial v_i}
\log p(y|x,v) 
= -\frac{X_j X_i}{\sigma^2},
\end{align*}
we have
\begin{eqnarray*}
I_{ij}
=\frac{J_{ij}}{\sigma^2}.
\end{eqnarray*}
Thus, the eigenvectors of $J$ coincide with those of the FIM $I$.
In practice, the matrix $I$ is approximated by the empirical FIM 
\begin{eqnarray*}
I^{(n)}_{ij}
= -\frac{1}{n}\sum_{t=1}^{n}\frac{\partial^2}{\partial v_i \partial v_j}\log p(x_{(t)},y_{(t)}|v).
\end{eqnarray*}
Define
\begin{eqnarray*}
J^{(n)}=\frac{1}{n}\sum_{t=1}^{n} {X_{(t)}}^T X_{(t)},
\end{eqnarray*}
where $X_{(1)},\ldots, X_{(n)}$ are $n$ independent realizations of $X$.
Then we have
\[
I^{(n)}=\frac{J^{(n)}}{\sigma^2}.
\]

\section{Main result}
\label{mainres}
In this section, we discuss the eigenvectors and eigenvalues of $J$. 

\subsection{Notation}
\label{prelms}
Hereafter, $W_l \in \Re^p$ denotes the $l$-th row vector of $W$, and 
$W^{(i)}\in \Re^d$ the $i$-th column vector of $W$.
For a vector $a$, $\|a\|$ denotes its Euclidean norm.
Note that $\|W^{(i)}\|\neq 0$ for all $i$ according to the assumption stated in Section \ref{onehid}.

Using the components of $W$, define the row vector $v^{(0)}\in \Re^p$ whose $i$-th component is
\begin{eqnarray*}
v^{(0)}_i=\frac{\|W^{(i)}\|}{\sqrt{d}}.
\end{eqnarray*}
We also define the row vector $v^{(\alpha,\beta)}\in \Re^p$ whose $i$-th component is  
\begin{eqnarray*}
v^{(\alpha,\beta)}_i= \frac{\sqrt{d}W_{\alpha i}W_{\beta i}}{\|W^{(i)}\|} \ (1\leq \alpha \leq \beta \leq d).
\end{eqnarray*}
Further, define the row vector $v^{(\gamma)}\in \Re^p$ by 
\begin{eqnarray*}
v^{(\gamma)}= \frac{v^{(\gamma,\gamma)}-v^{(0)}}{\sqrt{2}} \ (1\leq \gamma \leq d).
\end{eqnarray*}
Note that 
\begin{eqnarray}
\sum_{\gamma=1}^d v^{(\gamma)}= {\bf 0}
\label{sumgamma}
\end{eqnarray}
holds, since
\begin{eqnarray*}
\sum_{\gamma=1}^d v^{(\gamma,\gamma)}_i= \sum_{\gamma=1}^d\frac{\sqrt{d}W_{\gamma i}^2}{\|W^{(i)}\|}
=\frac{\sqrt{d}\|W^{(i)}\|^2}{\|W^{(i)}\|}=dv^{(0)}_i.
\end{eqnarray*}

\subsection{Summary of Findings}
\label{summary}
Suppose that each $W_{ij}$ is an independent realization of a random variable according to 
$N(0,1/p)$,
where $N(0,1/p)$ denotes the Gaussian distribution with mean 0 and variance $1/p$.
When $p$ is sufficiently large, the following holds for $J$ with high probability:
\begin{enumerate}
  \item The maximum eigenvalue is close to $(2d+1)/4\pi$, and the corresponding eigenvector is close to $v^{(0)}$.
  \item The second to $(d+1)$th largest eigenvalues are close to $1/4$, and the corresponding eigenvectors are close to $W_1, W_2, \ldots, W_d$.
  \item In the group of the next largest eigenvalues, the number of eigenvalues is $d(d+1)/2-1$ and the eigenvalues are close to $1/(2\pi d)$.
        The eigenspace of this group is close to the space spanned by $d(d+1)/2$ vectors $v^{(\alpha,\beta)}$ $(1\leq \alpha \leq \beta \leq d)$. 
\end{enumerate}
We formulate the above in Theorem \ref{thm_main} in Section \ref{eva_eigen}.

{\it Remark 1}: We can see that $p$ must be greater than $d(d+3)/2$	
in order to see the grouping of eigenvalues.	
Further, in Section \ref{eva_eigen},
we discuss the conditions for $p$ and $d$ for the above estimates of eigenvalues and eigenvectors to be accurate.	

{\it Remark 2}: If the activation function is the identity, that is, $X=xW$,	
we have only $d$ eigenvectors close to $W_1, W_2, \ldots, W_d$ whose	
eigenvalues are close to $1$.	
The eigenspaces of 1) and 3) in our case are due to the non-linearity of ReLU.	

{\it Remark 3}: The vector $v^{(0)}$ is the positive vector and	
approaches the vector $(1,\ldots ,1)/\sqrt{p}$ when $d$ becomes large.	
The vector $ v^{(0)} $ corresponds to the Perron-Frobenius eigenvector of the non-negative matrix $J$.	

{\it Remark 4}: When approximating $\|W^{(i)}\|$ by $\sqrt{d/p}$ for all $i$, 
we can express the eigenvectors in 3) as 	
\begin{eqnarray*}
v^{(\alpha,\beta)}\fallingdotseq \sqrt{p}\, W_\alpha \odot W_\beta,
\end{eqnarray*}
where $\odot$ denotes the Hadamard product.
This may be important in understanding the nature of eigenvectors in 3).

\subsection{Calculation of the matrix $J$}
In fact, each element of the matrix $J$ can be calculated and can be expressed using $W$
according to Theorem 1 in \cite{Tian} by Tian.
He investigated theoretical properties of 
a two-layer network with ReLU when the input $x$ follows a multivariate standard normal distribution.
The theorem is quoted below for the case that the sample size equals 1. 
\begin{thm}[Tian 2017]
\label{thm_tian}
Denote $F(e,w)=x^T D(e)D(w)x\cdot w$, where $e,w,x\in \Re^d$ are column vectors,
$e$ is the unit vector, 
and 
\begin{eqnarray*}
D(w) = 
\begin{cases}
        1   & \ \text{if $x\cdot w > 0$}\\
        0   & \ \text{if $x\cdot w \leq 0$.}
\end{cases}
\end{eqnarray*}
If $x$ is subject to the $d$-dimensional standard normal distribution (and thus bias-free), then
\begin{eqnarray*}
E[F(e,w)] = \frac{1}{2\pi}[(\pi-\theta)w^T + (\|w\|\sin\theta) e^T]
\end{eqnarray*}
where $\theta \in [0, \pi]$ is the angle between $e$ and $w$.
\end{thm}
\stepcounter{pft}

From Theorem \ref{thm_tian} with  $e=W^{(i)T}/\|W^{(i)}\|$ and $w=W^{(j)T}$, 
the following lemma is obtained:
\begin{lem}
\label{thm_fim}
Let $\theta_{ij}$ be the angle between non-zero vectors $W^{(i)}$ and $W^{(j)}$ ($0\leq \theta_{ij}\leq \pi$).
Then by Theorem 1, we have 
\begin{eqnarray*}
J_{ij}=\frac{1}{2\pi}((\pi -\theta_{ij})\cos \theta_{ij}+ \sin\theta_{ij})\|W^{(i)}\|\|W^{(j)}\|.
\end{eqnarray*}
In particular for $i=j$,
\begin{eqnarray*}
J_{ii}=\frac{1}{2}\|W^{(i)}\|^2,
\end{eqnarray*}
since $\theta_{ii}=0$.
\end{lem}

\begin{pf}
Calculating the quantity $J_{ij}=E[X_iX_j]$, we have
\begin{eqnarray*}
J_{ij}\!\!&=&\!\!E\left[\varphi\left(x\cdot W^{(i)}\right)\varphi\left(x \cdot W^{(j)}\right)\right]\\
\!\!&=&\!\!W^{(i)T}E\left[x^T D(W^{(i)}) D(W^{(j)}) x \cdot W^{(j)}\right]\\
\!\!&=&\!\!W^{(i)T}E\left[F\left(\frac{W^{(i)T}}{\|W^{(i)}\|},W^{(j)T}\right)\right]\\
\!\!&=&\!\! \frac{W^{(i)T}}{2\pi}\left[(\pi-\theta_{ij})W^{(j)} + (\|W^{(j)}\|\sin\theta_{ij}) \frac{W^{(i)}}{\|W^{(i)}\|}\right]\\
\!\!&=&\!\!\frac{1}{2\pi}\left[(\pi-\theta_{ij})W^{(i)}\cdot W^{(j)} + \|W^{(i)}\|\|W^{(j)}\|\sin\theta_{ij}\right]
\end{eqnarray*}
which yields the claim of the lemma.

\end{pf}

Further, $J_{ij}$ can be expanded as a polynomial for the inner product 
$W^{(i)} \cdot W^{(j)}$ as shown in the following lemma, 
which is useful for eigenvalue analysis of $J$.
\begin{lem}
\label{lem_fim}
Assume that all the columns of matrix $W$ are non-zero vectors.
For all $i$ and $j$, we have 
\begin{eqnarray*}
J_{ij}&=&\frac{A_{ij}}{2\pi}+\frac{W^{(i)}\! \cdot \!W^{(j)}}{4}+\frac{(W^{(i)}\!\cdot \!W^{(j)})^2}{4\pi A_{ij}}\\
&&+\frac{1}{2\pi}\sum_{n=1}^{\infty}\binom{2n}{n}\frac{(W^{(i)}\!\cdot \!W^{(j)})^{2n+2}}{2^{2n}(2n+1)(2n+2)A_{ij}^{2n+1}},
\end{eqnarray*}
where $A_{ij}=\|W^{(i)}\|\|W^{(j)}\|$.
\end{lem}

\begin{pf}
Let $\theta_{ij}$ be the angle between  $W^{(i)}$ and $W^{(j)}$ ($0\leq \theta_{ij}\leq \pi$), 
and $\bar{\theta}_{ij}=\pi/2-\theta_{ij}$ ($-\pi/2\leq \bar{\theta}_{ij}\leq \pi/2$).
Define $A_{ij}=\|W^{(i)}\|\|W^{(j)}\|$.
According to Lemma \ref{thm_fim}, it follows that 
\begin{eqnarray}
J_{ij}&=&\frac{A_{ij}}{2\pi}((\pi -\theta_{ij})\cos \theta_{ij}+ \sin\theta_{ij})\nonumber\\
&=&\frac{A_{ij}}{2\pi}(\bar{\theta}_{ij}\sin \bar{\theta}_{ij}+ \cos \bar{\theta}_{ij})+\frac{A_{ij}\cos \theta_{ij}}{4}\nonumber\\
&=&\frac{A_{ij}}{2\pi}f(\sin \bar{\theta}_{ij})+\frac{W^{(i)}\! \cdot \!W^{(j)}}{4},\label{Jij}
\end{eqnarray}
where we have defined $f(z)$ for $|z|\leq 1$
\begin{eqnarray*}
f(z)=z \arcsin z + \sqrt{1-z^2}.
\end{eqnarray*}
At the point of $|z|=1$, the first and second terms of $f (z)$ are not differentiable, but $f(z)$ as a whole is differentiable.
Then, the derivative $f'(z)$ is $\arcsin z$, which is expanded for $|z| \leq 1$ as the power series 
\begin{eqnarray*}
f'(z)=\sum_{n=0}^{\infty}\binom{2n}{n}\frac{z^{2n+1}}{2^{2n}(2n+1)}.
\end{eqnarray*}
(See \cite{Grad} for example.)
Since all the terms in the above series are positive for $z> 0$,
the convergence of the series for $z \in [0,1]$ implies the absolute convergence for $|z| \le 1$.
Thus, by the monotone convergence theorem we can integrate it term by term in any section within $[-1,1]$.
Noting that $f(0)=1$, we have
\begin{eqnarray}
f(z)=1+\sum_{n=0}^{\infty}\binom{2n}{n}\frac{z^{2n+2}}{2^{2n}(2n+1)(2n+2)}.\label{fz}
\end{eqnarray}
From (\ref{Jij}) and (\ref{fz}), and $\sin \bar{\theta}_{ij}=W^{(i)}\! \cdot \!W^{(j)}/A_{ij}$, we have
\begin{align*}
J_{ij} &= \frac{A_{ij}}{2\pi}\left(1+\sum_{n=0}^{\infty}\binom{2n}{n}
\frac{(W^{(i)} \cdot \!W^{(j)})^{2n+2}}{2^{2n}(2n+1)(2n+2)A_{ij}^{2n+2}}\right)\\
& +\frac{W^{(i)}\! \cdot \!W^{(j)}}{4}\\
&=\frac{A_{ij}}{2\pi}
+\frac{W^{(i)}\! \cdot \!W^{(j)}}{4}
+\frac{(W^{(i)}\!\cdot \!W^{(j)})^2}{4\pi A_{ij}}\\
&+\frac{1}{2\pi}\sum_{n=1}^{\infty}\binom{2n}{n}\frac{(W^{(i)}\!\cdot \!W^{(j)})^{2n+2}}{2^{2n}(2n+1)(2n+2)A_{ij}^{2n+1}},
\end{align*}
which completes the proof.
\end{pf}

\subsection{Decomposition of the matrix $J$}
As shown in the following theorem, the matrix $J$ can be expressed in the form of matrix sum
using the vectors defined in Section \ref{prelms}.
\begin{thm}
\label{lem_decomp}
Assume that
all the columns of matrix $W$ are non-zero vectors.
For the matrix $J=E[X^T X]$ and the vectors $W_l$ $(1\leq l \leq d)$, $v^{(0)}$, 
$v^{(\alpha,\beta)}$ $(1\leq \alpha < \beta \leq d)$, and $v^{(\gamma)}$  $(1\leq \gamma \leq d)$
defined in Section \ref{prelms}, we have
\begin{align}\label{jtheorem2}
J &= \frac{2d+1}{4\pi}v^{(0)T}v^{(0)}+\frac{1}{4}\sum_{l=1}^d W_l^T W_l \\ \nonumber
&+\frac{1}{2\pi d} \left(\sum_{\gamma=1}^d \!v^{(\gamma)T}v^{(\gamma)}
+\sum_{\alpha<\beta}\! v^{(\alpha,\beta)T}v^{(\alpha,\beta)}\!\right)+R,
\end{align}
where the matrix $R$, which is defined as
\begin{eqnarray*}
R_{ij}=\frac{1}{2\pi}\sum_{n=1}^{\infty}\binom{2n}{n}\frac{(W^{(i)}\!\cdot \!W^{(j)})^{2n+2}}{2^{2n}(2n+1)(2n+2)A_{ij}^{2n+1}},
\end{eqnarray*}
is positive-semidefinite.
\end{thm}

{\it Remark:}
If the vectors $v^{(0)}$, $W_l$, $v^{(\alpha,\beta)}$, and $v^{(\gamma)}$ 
are nearly orthonormal, and if $R$ is negligible,
\eqref{jtheorem2} can be regarded as an approximate
spectral decomposition of the FIM.
Intuitively, if each component of $W$ is independently generated from an appropriate distribution with mean 0, it is likely that the vectors $W_l$ are nearly orthonormal with high probability.
Further, we can expect that
the vectors $W_l$ and $v^{(\alpha,\beta)}$ are nearly orthogonal 
to each other with a high probability.

\

In the rest of this paper, 
we will show that the conditions stated in the above remark hold with a high probability
under the natural condition we assume.

\begin{pft}
To obtain the decomposition of the matrix $J$, 
we will find a matrix whose ($i$,$j$) component is equal to each term of the expansion of $J_{ij}$ in Lemma \ref{lem_fim}.
As for the first and the second terms, we can easily see that
\begin{eqnarray*}
A_{ij}=\|W^{(i)}\|\|W^{(j)}\|=d\left(v^{(0)T}v^{(0)}\right)_{ij}
\end{eqnarray*}
and
\begin{eqnarray*}
W^{(i)}\! \cdot \!W^{(j)}=\sum_{l=1}^d W_{li}W_{lj}=\sum_{l=1}^d (W_l^T W_l)_{ij}.
\end{eqnarray*}
For the third term,
expanding the square of the sum for $(W^{(i)}\!\cdot \!W^{(j)})^2$, we have
\begin{eqnarray*}
&&\!\!\!\!\!\!\!\!\!\!\!\!\frac{(W^{(i)}\!\!\cdot \!W^{(j)})^2}{A_{ij}}\\
&=&\!\!\!\!\!
\frac{\left(\sum_{\gamma=1}^d W_{\gamma i}W_{\gamma j}\right)^2}{\|W^{(i)}\|\|W^{(j)}\|}\\
&=&\!\!\!\!\!\sum_{\gamma=1}^d \frac{W_{\gamma i}^2 W_{\gamma j}^2}{\|W^{(i)}\|\|W^{(j)}\|}
\!+\!2\sum_{\alpha<\beta} \frac{W_{\alpha i}W_{\alpha j}W_{\beta i}W_{\beta j}}{\|W^{(i)}\|\|W^{(j)}\|}\\
&=&\!\!\frac{1}{d}\sum_{\gamma=1}^d \!\left(v^{(\gamma,\gamma)T}v^{(\gamma,\gamma)}\right)_{ij}
+\frac{2}{d}\!\sum_{\alpha<\beta}\! \left(v^{(\alpha,\beta)T}v^{(\alpha,\beta)}\right)_{ij},
\end{eqnarray*}
where the summation "${\alpha<\beta}$" takes for all $\alpha,\beta \in \{1,2,\ldots, d\}$ with $\alpha < \beta$.
The first term of the above is calculated as
\begin{eqnarray*}
&&\!\!\!\!\frac{1}{d}\sum_{\gamma=1}^d \!\left(v^{(\gamma,\gamma)T}v^{(\gamma,\gamma)}\right)_{ij}\\
&=&\frac{1}{d}\sum_{\gamma=1}^d \!\left((v^{(0)}+\sqrt{2}v^{(\gamma)T}) (v^{(0)}+\sqrt{2}v^{(\gamma)})\right)_{ij}\\
&=&\frac{1}{d}\sum_{\gamma=1}^d \!\left(v^{(0)T}v^{(0)}+2v^{(\gamma)T}v^{(\gamma)}+2\sqrt{2}v^{(\gamma)T}v^{(0)} \right)_{ij}
.
\end{eqnarray*}
Note that
\begin{eqnarray*}
\left(\sum_{\gamma=1}^d v^{(\gamma)T}v^{(0)}\right)_{ij}=\left({\bf 0}^T v^{(0)}\right)_{ij}=0
\end{eqnarray*}
from (\ref{sumgamma}).
Thus, the third term is
\begin{eqnarray*}
\frac{(W^{(i)}\!\!\cdot \!W^{(j)})^2}{A_{ij}}
&=&\!\!\left(v^{(0)T}v^{(0)}\right)_{ij}+\frac{2}{d}\sum_{\gamma=1}^d \!\left(v^{(\gamma)T}v^{(\gamma)}\right)_{ij}\\
&&+\frac{2}{d}\sum_{\alpha<\beta}\! \left(v^{(\alpha,\beta)T}v^{(\alpha,\beta)}\right)_{ij}.
\end{eqnarray*}

Plugging in the above terms to the formula in Lemma \ref{lem_fim}, the matrix $J$ is decomposed as
(\ref{jtheorem2}).

Here, we will show that the matrix $R$ is positive-semidefinite. 
The matrix $R$ is represented as
\begin{eqnarray*}
R=\frac{1}{2\pi}\sum_{n=1}^{\infty}\binom{2n}{n}\frac{1}{2^{2n}(2n+1)(2n+2)}R^{(n)},
\end{eqnarray*}
where we have defined the matrix $R^{(n)}$ as
\begin{eqnarray*}
R^{(n)}_{ij}&=&\frac{(W^{(i)}\!\cdot \!W^{(j)})^{2n+2}}{A_{ij}^{2n+1}}\\
&=&\left(\frac{W^{(i)}}{\|W^{(i)}\|}\!\cdot \!\frac{W^{(j)}}{\|W^{(j)}\|}\right)^{2n+2}\|W^{(i)}\|\|W^{(j)}\|.
\end{eqnarray*}
As shown below, $R^{(n)}$ is positive-semidefinite for all $n\geq 1$, which completes the proof.
Let $G$ be the $p\times p$ matrix whose $(i, j)$ component is $(W^{(i)}/\|W^{(i)}\|)\!\cdot \!(W^{(j)}/\|W^{(j)}\|)$.
Then $G$ is a Gram matrix of the set of vectors $\{W^{(i)}/\|W^{(i)}\||i=1,\ldots,p\}$, and 
is positive-semidefinite.
Letting $\odot$ denote the Hadamard product, we can represent $R^{(n)}$ as 
\begin{eqnarray*}
R^{(n)}=\overbrace{G \odot G \odot \cdots \odot G}^{2n+2} \odot (dv^{(0)T}v^{(0)}).
\end{eqnarray*}
Thus, $R^{(n)}$ is positive-semidefinite by Schur Product Theorem (see \cite{Bapat} for example).


\end{pft}


\subsection{Evaluation of eigenvalues}
\label{eva_eigen}
Recall that $W_{ij}$s are realizations of the independent random variables drawn from $N(0,1/p)$.
When $p$ is large, the norms of the vectors $W_l$ $(1\leq l \leq d)$, $v^{(0)}$, 
$v^{(\alpha,\beta)}$ $(1\leq \alpha < \beta \leq d)$, and $v^{(\gamma)}$  $(1\leq \gamma \leq d)$
are almost equal to 1 with high probability.
Further, the vectors are almost orthogonal to each other except for $v^{(\gamma)}$ and $v^{(\gamma')}$ ($\gamma\neq \gamma'$).
From equation (\ref{sumgamma}), the vectors $v^{(\gamma)}$  $(1\leq \gamma \leq d)$  are linearly dependent.
However, we can take $d-1$ orthonormal basis from the vectors $v^{(1)},v^{(2)},\ldots,v^{(d)}$.

We formulate the above as the following lemma
.
To do so, we introduce a quantity $\xi(d)$, which converges to 0 as $d$ goes to infinity, where $\xi(d)=O(1/d^{1/2-\eta})$ for arbitrary $\eta\in(0,1/2)$.
The definition of $\xi(d)$ is given in Appendix \ref{defxi}.

\begin{lem}
\label{lem_orthnorm}
Let each $W_{ij}$ be realization of the independent random variable drawn from $N(0,1/p)$, 
let $C$ be a certain positive constant that does not depend on $p$ and $d$, $D=(d+1)(d+2)(d^2+3d+4)/8$, and $d>4$.
Then for all $\delta>0$, the following holds with the probability $1-CD/(\delta^2p)$ at least:
\begin{enumerate}
  \item For the vectors $v^{(0)}$, $W_l$ $(1\leq l \leq d)$,
  $v^{(\alpha,\beta)}$ $(1\leq \alpha < \beta \leq d)$, and $v^{(\gamma)}$  $(1\leq \gamma \leq d)$
  defined in Section \ref{prelms}, we have
  \begin{eqnarray*}
  |\|v^{(0)}\|^2-1| &\leq& \delta,\\
  |\|W_l\|^2-1| &\leq& \delta,\\
  |\|v^{(\alpha,\beta)}\|^2-1| &\leq& \delta+\xi(d),\\
  |\|v^{(\gamma)}\|^2-1| &\leq& \delta+\xi(d).
  \end{eqnarray*}
 
 \item Let $V_1=\{v^{(0)},W_l,v^{(\alpha,\beta)}|1\leq l \leq d,1\leq \alpha < \beta \leq d\}$,
       $V_2=\{v^{(\gamma)}|1\leq \gamma \leq d\}$, and $V=V_1\cup V_2$.
       We have for all different vectors $v,v'\in V$ ($v\notin V_2$ or $v'\notin V_2$)
       \begin{eqnarray*}
       |v\cdot v'|\leq \delta,
       \end{eqnarray*}
       and for all different vectors $v,v'\in V_2$
       \begin{eqnarray*}
       \left|v\cdot v' -\left(-\frac{1}{d-1}\right)\right|\leq \delta+\frac{\xi(d)}{d-1}.
       \end{eqnarray*}
\end{enumerate}
\end{lem}
Proof of this lemma is stated in next subsection.

From this lemma, 
we can see that Theorem \ref{lem_decomp} gives the eigenvalue decomposition of $J$ approximately if $R$ can be ignored.
In fact, the following theorem shows that the effect of $R$ on the eigenvalues of $J$ is small.

\begin{thm}
\label{thm_main}
Let each $W_{ij}$ be realization of the independent random variable drawn from $N(0,1/p)$, 
let $C$ be a certain positive constant that does not depend on $p$ and $d$, $D=(d+1)(d+2)(d^2+3d+4)/8$, and $d>4$.
Then for all $\delta>0$, the following holds with the probability of $1-CD/(\delta^2p)$ at least:
\begin{enumerate}
  \item The sum of the eigenvalues of the matrix $J$ is bounded as
  \begin{eqnarray*}
  \mathrm{tr}(J)\leq \frac{d}{2}(1+\delta).
  \end{eqnarray*}
  \item For the vectors $v^{(0)}$, $W_l$ $(1\leq l \leq d)$,
  $v^{(\alpha,\beta)}$ $(1\leq \alpha < \beta \leq d)$, and $v^{(\gamma)}$  $(1\leq \gamma \leq d)$
  defined in Section \ref{prelms}, we have
 \begin{eqnarray*}
 \frac{v^{(0)}}{\|v^{(0)}\|}J \frac{v^{(0)T}}{\|v^{(0)}\|}&\geq& \frac{2d+1}{4\pi}(1-\delta),\\
 \frac{W_{l}}{\|W_{l}\|}J \frac{W_{l}^T}{\|W_{l}\|}&\geq& \frac{1}{4}(1-\delta),\\
 \frac{v^{(\alpha,\beta)}}{\|v^{(\alpha,\beta)}\|}J \frac{v^{(\alpha,\beta)T}}{\|v^{(\alpha,\beta)}\|}&\geq& 
 \frac{1}{2\pi d}\left(1-\delta-\xi(d)\right),\\
 \frac{v^{(\gamma)}}{\|v^{(\gamma)}\|}J \frac{v^{(\gamma)T}}{\|v^{(\gamma)}\|}&\geq& \frac{1}{2\pi d}
 \left(1-\delta-\xi(d)\right).
 \end{eqnarray*}
\end{enumerate}
\end{thm}


We give an interpretation of Lemma \ref{lem_orthnorm} and Theorem \ref{thm_main} below. 
Taking a sufficiently large $p/D$ for an arbitrarily small $\delta$, the events 
in the lemma and the theorem hold with high probability.
The inequalities in Lemma \ref{lem_orthnorm} show that the different vectors belonging to $V$ are almost orthonormal to each other
except for the vectors belonging to $V_2$.
The inequalities in 2) of Theorem \ref{thm_main} give a lower bound of the amount corresponding to the eigenvalue for the direction of each vector belonging to $V$.	
Hence, we can estimate the sum of all the eigenvalues related to $V$ from the
sum of the right-hand sides of the inequalities in 2) for each vector.
Ignoring $\delta$ and $\xi(d)$, and noting that the space spanned by $v^{(\gamma)}$s is	
$d-1$ dimensional, the sum is
\begin{eqnarray*}
&&\frac{2d+1}{4\pi}+\frac{1}{4}\cdot d+\frac{1}{2\pi d}\cdot\left(\frac{d(d-1)}{2}+d-1\right)\\
&\geq&\frac{3+\pi}{2\pi}\cdot \frac{d}{2}\geq 0.977\cdot \frac{d}{2},
\end{eqnarray*}
which occupies most of the upper bound on the trace of $J$ in 1).

\begin{pft}
\begin{enumerate}
\item The trace of the matrix $J$ is
  \begin{eqnarray*}
  \mathrm{tr}(J)= \sum_{i=1}^p J_{ii}= \frac{1}{2}\sum_{i=1}^p \|W^{(i)}\|^2 =\frac{1}{2}\sum_{l=1}^d \|W_l\|^2.
  \end{eqnarray*}
  From Lemma \ref{lem_orthnorm}, we have for all $1\leq l \leq d$
  \begin{eqnarray*}
  \left|\|W_l\|^2-1\right| \leq \delta,
  \end{eqnarray*}
  with the probability of $1-CD/(\delta^2p)$ at least. Then, $\mathrm{tr}(J)$ satisfies
  \begin{eqnarray*}
  \mathrm{tr}(J)\leq \frac{d}{2}(1+\delta).
  \end{eqnarray*}

\item From Theorem \ref{lem_decomp}, we have 
  \begin{eqnarray*}
  \frac{v^{(0)}}{\|v^{(0)}\|}J \frac{v^{(0)T}}{\|v^{(0)}\|}&\geq& \frac{2d+1}{4\pi}\|v^{(0)}\|^2,\\
  \frac{W_{l}}{\|W_{l}\|}J \frac{W_{l}^T}{\|W_{l}\|}&\geq& \frac{1}{4}\|W_{l}\|^2,\\
  \frac{v^{(\alpha,\beta)}}{\|v^{(\alpha,\beta)}\|}J \frac{v^{(\alpha,\beta)T}}{\|v^{(\alpha,\beta)}\|}
  &\geq& \frac{1}{2\pi d}\|v^{(\alpha,\beta)}\|^2,\\
  \frac{v^{(\gamma)}}{\|v^{(\gamma)}\|}J \frac{v^{(\gamma)T}}{\|v^{(\gamma)}\|}&\geq& \frac{1}{2\pi d}\|v^{(\gamma)}\|^2.
  \end{eqnarray*}
  Further from Lemma \ref{lem_orthnorm},
  we have for all $1\leq l \leq d$, $1\leq \alpha < \beta \leq d$, and $1\leq \gamma \leq d$
  \begin{eqnarray*}
  \|v^{(0)}\|^2 &\geq& 1-\delta,\\
  \|W_l\|^2 &\geq& 1-\delta,\\
  \|v^{(\alpha,\beta)}\|^2 &\geq& 1-\delta-\xi(d),\\
  \|v^{(\gamma)}\|^2 &\geq& 1-\delta-\xi(d)
  \end{eqnarray*}
  with the probability of $1-CD/(\delta^2p)$ at least. 
  Then, we have 
  \begin{eqnarray*}
  \frac{v^{(0)}}{\|v^{(0)}\|}J \frac{v^{(0)T}}{\|v^{(0)}\|}&\geq& \frac{2d+1}{4\pi}(1-\delta),\\
  \frac{W_{l}}{\|W_{l}\|}J \frac{W_{l}^T}{\|W_{l}\|}&\geq& \frac{1}{4}(1-\delta),\\
  \frac{v^{(\alpha,\beta)}}{\|v^{(\alpha,\beta)}\|}J \frac{v^{(\alpha,\beta)T}}{\|v^{(\alpha,\beta)}\|},&\geq& 
  \frac{1}{2\pi d}\left(1-\delta-\xi(d)\right),\\
  \frac{v^{(\gamma)}}{\|v^{(\gamma)}\|}J \frac{v^{(\gamma)T}}{\|v^{(\gamma)}\|}&\geq& \frac{1}{2\pi d}
  \left(1-\delta-\xi(d)\right).
  \end{eqnarray*}
\end{enumerate}
The proof is completed by 1) and 2) above.
\end{pft}

\subsection{Proof of Lemma \ref{lem_orthnorm}}
\label{sec_orthnorm}
To prove Lemma \ref{lem_orthnorm}, we use the following lemma 
that evaluates the expectation and variance for the norms and inner products of the vectors.
\begin{lem}
\label{lem_exp_var}
Let each $W_{ij}$ be realization of the independent random variable drawn from $N(0,1/p)$, 
let $C$ be a certain positive constant that does not depend on $p$ and $d$, and $d>4$.
Then, the following holds.
\begin{enumerate}
  \item For the vectors $v^{(0)}$, $W_l$ $(1\leq l \leq d)$,
  $v^{(\alpha,\beta)}$ $(1\leq \alpha < \beta \leq d)$, and $v^{(\gamma)}$  $(1\leq \gamma \leq d)$
  defined in Section \ref{prelms}, we have
  \begin{eqnarray*}
  E[\|v^{(0)}\|^2] &=& 1,\\
  E[\|W_l\|^2] &=& 1,\\
  |E[\|v^{(\alpha,\beta)}\|^2]-1| &\leq& \xi(d),\\
  |E[\|v^{(\gamma)}\|^2]-1| &\leq& \xi(d).
  \end{eqnarray*}
 
 \item Let $V_1=\{v^{(0)},W_l,v^{(\alpha,\beta)}|1\leq l \leq d,1\leq \alpha < \beta \leq d\}$,
       $V_2=\{v^{(\gamma)}|1\leq \gamma \leq d\}$, and $V=V_1\cup V_2$.
       We have for all different vectors $v,v'\in V$ ($v\notin V_2$ or $v'\notin V_2$)
       \begin{eqnarray*}
       E[v\cdot v']=0,
       \end{eqnarray*}
       and for all different vectors $v,v'\in V_2$
       \begin{eqnarray*}
       \left|E[v\cdot v'] -\left(-\frac{1}{d-1}\right)\right|\leq \frac{\xi(d)}{d-1}.
       \end{eqnarray*}
 \item For all vectors $v,v'\in V$, the variance of the inner product is bounded as
       \begin{eqnarray*}
       Var[v\cdot v']\leq \frac{C}{p}.
       \end{eqnarray*}
\end{enumerate}
\end{lem}
\stepcounter{pft}

Proof of this lemma is stated in Appendix \ref{orth}, and some lemmas we used to prove Lemma \ref{lem_exp_var} are given in Appendix \ref{exp_prob}, \ref{exp_upp}, and \ref{exp_low}.

\begin{pf}
Define the set of random variables $V_{pair}=\{v\cdot v'|v,v' \in V \}$.
Since the cardinality of $V$ is $d(d+3)/2+1$, that of $V_{pair}$ is 
\begin{eqnarray*}
&&\frac{1}{2}\left(\frac{d(d+3)}{2}+1\right)\left(\frac{d(d+3)}{2}+2\right)\\
&=&\frac{(d+1)(d+2)(d^2+3d+4)}{8}=D.
\end{eqnarray*}
From Lemma \ref{lem_exp_var}, it follows that $Var(T)\leq C/p$ for all $T \in V_{pair}$, where $C$ is a certain positive constant.
Thus, from the Chebyshev's inequality, we have for all $T \in V_{pair}$
\begin{eqnarray*}
\Pr(|T-E(T)|\geq \delta)\leq \frac{C}{\delta^2 p}.
\end{eqnarray*}
Further, using the union bound for the above, we have 
\begin{eqnarray}
&&\Pr(\exists T \in V_{pair},\ |T-E(T)|\geq \delta)\leq \frac{CD}{\delta^2 p}\nonumber\\
&\Leftrightarrow&\Pr(\forall T \in V_{pair},\ |T-E(T)|\leq \delta)\geq 1-\frac{CD}{\delta^2 p}.\label{event}
\end{eqnarray}
Thus, the following holds with the probability of $1-CD/(\delta^2p)$ at least:
\begin{enumerate}
  \item Considering the case of $T=\|v^{(0)}\|^2$ or $T=\|W_l\|^2$, we have $E(T)=1$ from Lemma \ref{lem_exp_var}.
        Since $|T-E(T)|\leq \delta$ holds for all $T \in V_{pair}$, we have
        \begin{eqnarray*}
        |\|v^{(0)}\|^2-1|&\leq& \delta,\\
        |\|W_l\|^2-1|&\leq& \delta.
        \end{eqnarray*}
        Next, considering the case of $T=\|v^{(\alpha,\beta)}\|^2$ or $T=\|v^{(\gamma)}\|^2$,
        we have $|E(T)-1| \leq \xi(d)$ from Lemma \ref{lem_exp_var}.
        Thus, we have 
        \begin{eqnarray*}
        |\|v^{(\alpha,\beta)}\|^2-1|&\leq& \delta+\xi(d),\\
        |\|v^{(\gamma)}\|^2-1|&\leq& \delta+\xi(d).
        \end{eqnarray*}

  \item 
  Considering the case $T=v\cdot v'$ for all different vectors $v,v'\in V$ ($v\notin V_2$ or $v'\notin V_2$),
  we have  $E(T) =0$ from Lemma \ref{lem_exp_var}. 
  Since $|T-E(T)|\leq \delta$ holds for all $T \in V_{pair}$, 
  we have for all different vectors $v,v'\in V$ ($v\notin V_2$ or $v'\notin V_2$)
  \begin{eqnarray*}
  |v\cdot v'-0| \leq \delta.
  \end{eqnarray*}
  Next, considering the case $T=v\cdot v'$ for all different vectors $v,v'\in V_2$,
  we have  
  \begin{eqnarray*}
  \left|E(T)-\left(-\frac{1}{d-1}\right)\right|\leq \frac{\xi(d)}{d-1}.
  \end{eqnarray*}
  from Lemma \ref{lem_exp_var}. 
  Thus, we have for all different vectors $v,v'\in V_2$
  \begin{eqnarray*}
  \left|v \cdot v'-\left(-\frac{1}{d-1}\right)\right|\leq \delta+\frac{\xi(d)}{d-1}.
  \end{eqnarray*}
\end{enumerate}
The proof is completed by 1) and 2) above.
\end{pf}

\section{Numerical Calculation}
\label{simulate}
\begin{figure}[t]
\centering
\includegraphics[width=8.5cm]{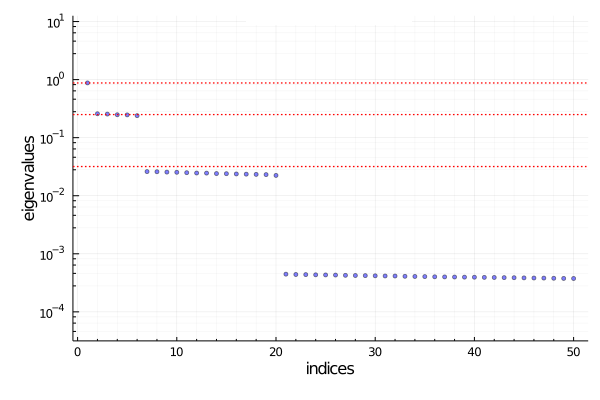}
\caption{Case of $d=5$}
\label{fig:d5}
\end{figure}

\begin{figure}[t]
\centering
\includegraphics[width=8.5cm]{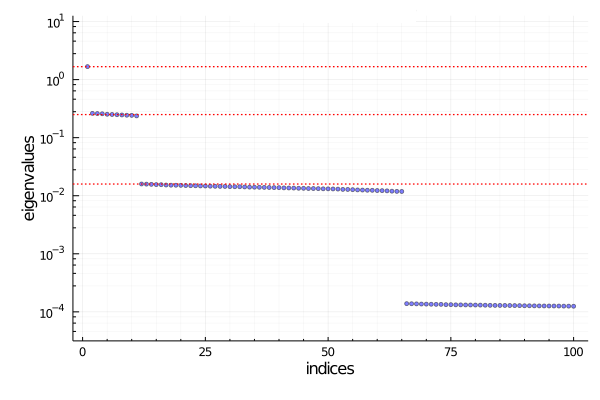}
\caption{Case of $d=10$}
\label{fig:d10}
\end{figure}

\begin{figure}[t]
\centering
\includegraphics[width=8.5cm]{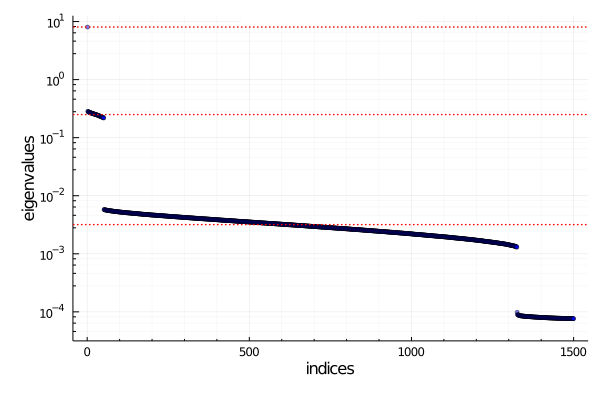}
\caption{Case of $d=50$}
\label{fig:d50}
\end{figure}

We computed the FIM of a one hidden layer network
in the setting of Theorem \ref{thm_main} and
obtained the eigenvalue decomposition by numerical calculation.
As a result, 
we confirmed that our estimates of the first three clusters of eigenvalues and eigenvectors shown in Theorem \ref{thm_main} are consistent with the numerical calculations.
We generated random numbers for $W$ as described in Section \ref{summary}, and calculated the matrix $J$ by using Lemma \ref{thm_fim}.
We set $p=10000$ and $d=5,10,20,50$, and calculated the eigenvalues of $J$ 
for each $d$. Fig.~\ref{fig:d5} to Fig.~\ref{fig:d50} show the magnitudes of the first three clusters of eigenvalues arranged in descending order, respectively.
For each $d$, we can see the grouping property of magnitudes of eigenvalues, and the number of each group 
matches with that stated in Section \ref{summary}.
The dotted lines drawn horizontally represent the values of $(2d+1)/4\pi$, $1/4$, $1/(2\pi d)$ in order from the top.
These values are approximations of the eigenvalues of each group from 2) of Theorem \ref{thm_main}.

\begin{figure}[t]
\centering
\includegraphics[width=8.5cm]{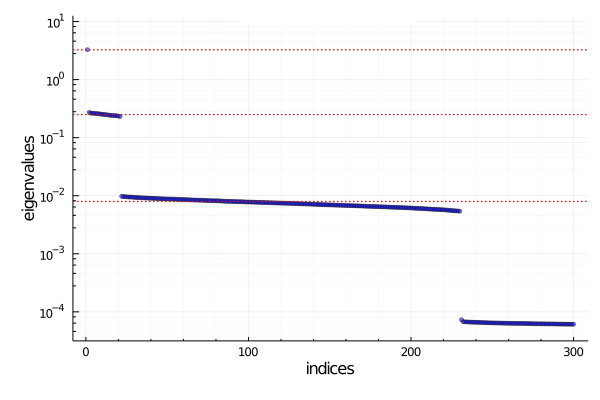}
\caption{Case of $d=20$}
\end{figure}
\label{fig:d20}

\begin{figure}[t]
\centering
\includegraphics[width=8.5cm]{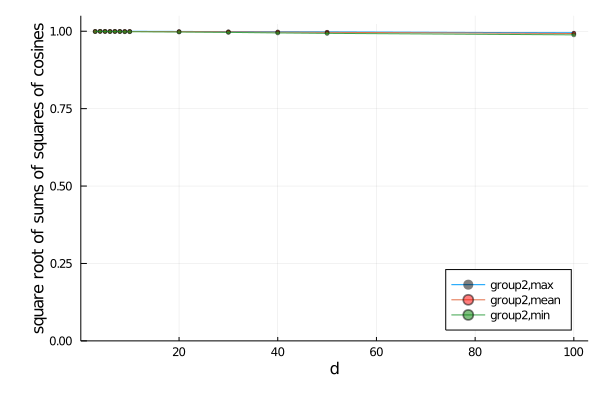}
\caption{The maximum, the mean, and the minimum of $q^{(\alpha,\beta)}$ for some $d$}
\label{fig:minq1}
\end{figure}

\begin{figure}[t]
\centering
\includegraphics[width=8.5cm]{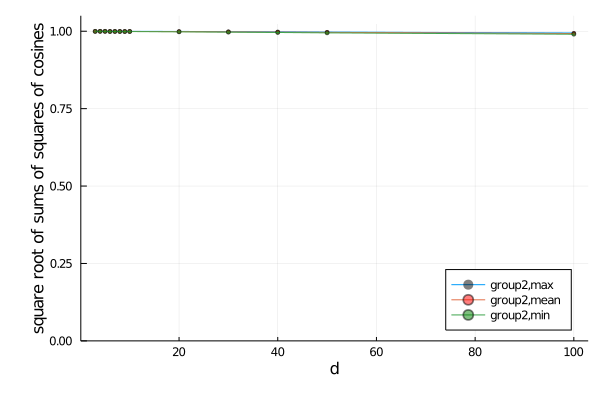}
\caption{The maximum, the mean, and the minimum of $q^{(\gamma)}$ for some $d$}
\label{fig:minq2}
\end{figure}

Next, we examined whether $v^{(\alpha,\beta)}$ and $v^{(\gamma)}$ are actually contained in the eigenspace of the third cluster obtained by numerical calculation.
Letting $\{u_k\}$ be an orthonormal basis of the eigenspace of the third cluster obtained numerically, we computed the following quantities;
\begin{align*}
q^{(\alpha,\beta)}=\sqrt{\sum_k 
\left(\frac{v^{(\alpha,\beta)}\cdot u_k}{\|v^{(\alpha,\beta)}\| \|u_k\|}
\right)^2}
\end{align*}
\begin{align*}
q^{(\gamma)}=\sqrt{\sum_k 
\left(\frac{v^{(\gamma)}\cdot u_k}{\|v^{(\gamma)}\| \|u_k\|}\right)^2}
\end{align*}
The closer $q^{(\alpha,\beta)}$ and $q^{(\gamma)}$ are to 1, the more components of $v^{(\alpha,\beta)}$ and $v^{(\gamma)}$ are contained in the eigenspace of the third cluster, respectively. For some values of $d$, the quantities $q^{(\alpha,\beta)}$ are examined in Fig.~\ref{fig:minq1} and $q^{(\gamma)}$ in Fig.~\ref{fig:minq2}. These graphs show that most components of $v^{(\alpha,\beta)}$ and $v^{(\gamma)}$ are contained in the eigenspace of the third cluster.

\section{Behavior of gradient descent}
\label{discuss}
In this section, we describe how the 
effective number in the learning network parameters by gradient descent is related to the 
eigenvalues of the FIM, based on Exercise 5.25 in \cite{PRML}.
The empirical loss for the $n$ training data 
$\left\{(x_{(i)},y_{(i)})|i=1,\ldots,n\right\}$ is defined as 
\begin{align*}
E(v)=\frac{1}{2n}\sum_{i=1}^n \left(y_{(i)}-X_{(i)}v^T\right)^2,
\end{align*}
where $X_{(i)}=\varphi(x_{(i)}W)$.
Here for a vector input, $\varphi$ is applied to each element.
Let $\hat{v}$ be the $v$ that minimizes $E(v)$.
Considering the Taylor expansion at $v=\hat{v}$, we have
\begin{align}
E(v)= E(\hat{v})+\frac{1}{2}(v-\hat{v})J^{(n)}(v-\hat{v})^T,\label{E_Taylor}
\end{align}
where $J^{(n)}$ is defined in Section \ref{FIMsec}, which is proportional to the
empirical FIM.
Note that $E(v)$ is quadratic with respect to $v$, so the third or higher order terms in the Taylor expansion do not appear.
Let us consider to learn the parameter $v$ by simple gradient descent
\begin{align}
v(\tau)=v(\tau-1)-\rho \nabla E(v(\tau-1)),
\label{gradient_ud}
\end{align}
where $v(\tau)$ denotes the value of $v$ after $\tau$-th update ($\tau=1,2,\ldots$) and $\rho$ is the learning rate (small positive number)
.
Let $u_j$ be the eigenvectors of $J^{(n)}$ and the corresponding eigenvalues $\eta_j$, that is,
$u_j J^{(n)}= \eta_j u_j$,
and $v_j(\tau)=v(\tau) u_j^T$
be the components of $v(\tau)$ parallel to $u_j$. 
Calculating the gradients of both sides of \eqref{E_Taylor}, we have
\begin{align*}
\nabla E(v(\tau-1))= (v(\tau-1)-\hat{v})J^{(n)}.
\end{align*}
Substituting the above into \eqref{gradient_ud}, we have
\begin{align*}
v(\tau)=v(\tau-1)-\rho (v(\tau-1)-\hat{v})J^{(n)}.
\end{align*}
Multiplying both sides with $u^T_j$ from the right, we have
\begin{align*}
v_j(\tau)=v_j(\tau-1)-\rho (v_j(\tau-1)-\hat{v}_j)\eta_j.
\end{align*}
Subtracting $\hat{v}_j$ from both sides, we have
\begin{align*}
v_j(\tau)-\hat{v}_j
=v_j(\tau-1)-\hat{v}_j-\rho \eta_j(v_j(\tau-1)-\hat{v}_j),
\end{align*}
which is
\begin{align*}
v_j(\tau)-\hat{v}_j
=(1-\rho \eta_j)(v_j(\tau-1)-\hat{v}_j).
\end{align*}
Suppose now that $\rho \eta_j << 1$. Then, we have
\[
v_j(\tau)-\hat{v}_j
 \fallingdotseq
\exp(-\rho \eta_j)(v_j(\tau-1)-\hat{v}_j),
\]
which yields
\[
(v_j(\tau)-\hat{v}_j
)^2 \fallingdotseq
\exp(-\tau/T_j)(v_j(0)-\hat{v}_j)^2.
\]
Here, we have defined $T_j=(2\rho \eta_j)^{-1}$.
The above formula means that
the training error for $v_j$ exponentially converges to $0$
with the time constant $T_j=(2\rho \eta_j)^{-1}$.
That is, if $\tau$ is larger enough than $T_j$,
the training error for $v_j$ is almost $0$,
while if $\tau$ is smaller than $T_j$,
it remains near the initial value.
Note that $\tau > T_j$ is equivalent to $\eta_j > (2\rho \tau)^{-1}$.

Thus, we can interpret the number of eigenvalues much larger than $(2\rho\tau)^{-1}$ as the effective number of parameters in the learning up to step $\tau$.
Now we suppose $J^{(n)}$ is so close to $J$
that it has the grouping property shown in this paper, which was valid 
in our experiment with $n > 10 d^2$.
Then, the grouping property 
implies that the effective number of parameters increases stepwise from $O(1)$, $O(d)$, $O(d^2)$ for raising the goal for accuracy of estimation.

Exercise 5.25 in \cite{PRML} discusses learning weights in neural networks in general, not just in the setting of this paper.
In this case, the third or higher order terms appear in the Taylor expansion of \eqref{E_Taylor}, but if we ignore the influence of those terms, we can make the same argument as above.
It is a future work to investigate whether the grouping of eigenvalues occurs in general neural networks.

The above grouping of eigenvalues results in a time lag between the end of learning for one group and the start of learning for the next group, which is known as a plateau.
The natural gradient \cite{Amari3},
which is the gradient times the inverse of the Fisher information,
is a well known remedy for plateaus.
If $J \fallingdotseq J^{(n)}$ in our case,
the natural gradient is almost equal to the direction of
$(\hat{v}-v)$,
then 
the plateau can be suppressed and we can effectively learn $v$ close to $\hat{v}$.
However, it means that we
does not take advantage of 
the structure of the effective parameters.
We may need to consider which method is preferable.
While we would not discuss it further in this paper, 
we may find that the simple gradient method is better than the natural gradient method in the view point of the overfitting problem.

\section{Conclusion}
\label{concl}
We approximately derived the main eigenvalues and eigenvectors of the Fisher information matrix 
of the one hidden layer neural network under certain conditions.
Specifically, we characterized the first three clusters of the eigenvalues and eigenvectors.
It is expected that this study will advance the investigation of the behavior of 
generalization error of deep learning, 
especially in the situation of overfitting mentioned in Section \ref{Related}.



\section*{Acknowledgment}
The authors give their sincere gratitude to
Professors Hiroshi Nagaoka, Noboru Murata, and Kazushi Mimura
for the valuable discussion with them.
This work was supported by JST SPRING, Grant Number JPMJSP2136.

\appendices

\section{Definition of $\xi(d)$}
\label{defxi}
For arbitrary $\eta\in(0,1/2)$, define 
\begin{eqnarray*}
\xi(d)=\max\left(\xi_1(d_1),\xi_2(d_2),\bar{\xi}_1(d_1),\bar{\xi}_1(d_2)\right),
\end{eqnarray*}
where we have defined $d_1=d-2$, $d_2=d-1$,
\begin{eqnarray*}
\xi_1(d_1)&=&1-(1-\iota_1(d_1^\eta))^2 
\frac{\left(1\!-\!2\exp\!\left(-\frac{d_1^{2\eta}}{8}\right)\right)(d_1+2)}{d_1+d_1^{1/2+\eta}+2d_1^{2\eta}},
\end{eqnarray*}
\begin{eqnarray*}
\xi_2(d_2)&=&\frac{3}{2}-\frac{3-\iota_2(d_2^\eta)}{2}
\frac{\left(1\!-\!2\exp\!\left(-\frac{d_2^{2\eta}}{8}\right)\right)(d_2+1)}{d_2+d_2^{1/2+\eta}+d_2^{2\eta}},
\end{eqnarray*}
\begin{eqnarray*}
\bar{\xi}_1(d_1)&=&\left(1+\frac{2}{d_1}\right) \left(1+\frac{1}{(d_1)^{1/2-\eta}-1}\right)\\
&&+2(d_1+2) \exp\left(-\frac{(d_1)^{2\eta}}{8}\right)-1,
\end{eqnarray*}
and
\begin{eqnarray*}
\bar{\xi}_2(d_2)&=&
\frac{3}{2}\left(1+\frac{1}{d_2}\right)\left(1+\frac{1}{(d_2)^{1/2-\eta}-1}\right)\nonumber\\
&&+2(d_2+1) \exp\left(-\frac{(d_2)^{2\eta}}{8}\right)-\frac{3}{2}.
\end{eqnarray*}
Here, we have defined the functions $\iota_1$ and $\iota_2$ as
\begin{eqnarray*}
\iota_1(a)=\sqrt{\frac{2}{\pi}}\left(a+\frac{1}{a}\right)\exp\left(-\frac{a^2}{2}\right)
\end{eqnarray*}
and
\begin{eqnarray*}
\iota_2(a)=\sqrt{\frac{2}{\pi}}\left(a^3+3a+\frac{3}{a}\right)\exp\left(-\frac{a^2}{2}\right).
\end{eqnarray*}
Noting that 
\begin{eqnarray*}
\frac{d_1+2}{d_1+d_1^{1/2+\eta}+2d_1^{2\eta}}
&=&1-\frac{d_1^{1/2+\eta}+2d_1^{2\eta}-2}{d_1+d_1^{1/2+\eta}+2d_1^{2\eta}}\\
&\leq&1-\frac{d_1^{1/2+\eta}+2d_1^{2\eta}-2}{4d_1}\\
&=&1-\frac{1}{4}\left(\frac{1}{d_1^{1/2-\eta}}+\frac{2}{d_1^{1-2\eta}}-\frac{2}{d_1}\right),\\
\end{eqnarray*}
we have
\begin{eqnarray*}
\xi_1(d_1)=O\left(\frac{1}{d_1^{1/2-\eta}}\right).
\end{eqnarray*}
Thus, we have
\begin{eqnarray*}
\xi(d)=O\left(\frac{1}{d^{1/2-\eta}}\right).
\end{eqnarray*}

\section{Tail probability}
\label{exp_prob}
Let $Z=(Z_1,\ldots,Z_d)$ denote a $d$-dimensional standard normal random vector.
Then, the following lemma holds. (see (2.19) in \cite{Wain}, for example)
\begin{lem}
\label{lem_prob}
Let $Z=(Z_1,\ldots,Z_d)$ denote a $d$-dimensional standard normal random vector.
Then for arbitrary $t\in(0,\sqrt{d})$, we have 
\begin{eqnarray*}
\Pr\left(\left|\frac{\|Z\|^2}{d}-1\right|\geq \frac{t}{\sqrt{d}}\right)\leq 2\exp\left(-\frac{t^2}{8}\right).
\end{eqnarray*}
\end{lem}
\stepcounter{pf}

\section{Upper bound on the expectation of some quantities}

\label{exp_upp}
We give upper bounds on the expectations of some random variables.
The upper bounds are shown in the following two lemmas.

\begin{lem}
\label{lem_upp}
Let $Z=(Z_1,\ldots,Z_d)$ denote a $d$-dimensional standard normal random vector.
Let $h$ be a natural number satisfying $h<d$ and 
$m_1,\ldots,m_h$ be natural numbers.
Further, define $d'=d-h$. Then for arbitrary $\kappa \in(0,1)$, we have 
\begin{align}
E\left[\frac{d\prod_{k=1}^h Z_k^{2m_k}}{\|Z\|^2} \right]
&\leq \frac{1}{\kappa} \left(1+\frac{h}{d'}\right) M_1 \label{upp_kappa} \\
&+2(d'+h)\exp\left(-\frac{(1-\kappa)^2d'}{8}\right) M_2, \nonumber
\end{align}
where we have defined $M_1$ and $M_2$ as
\begin{align*}
M_1 &=\prod_{k=1}^h E\left[ Z_k^{2m_k } \right], \\
M_2 &=E\left[ Z_1^{2(m_1-1) } \right] \prod_{k=2}^h E\left[ Z_k^{2m_k} \right].
\end{align*}
In particular, letting $1-\kappa=1/(d')^{1/2-\eta}$ for arbitrary $\eta\in(0,1/2)$, we have
\begin{align*}
E\left[\frac{d\prod_{k=1}^h Z_k^{2m_k}}{\|Z\|^2} \right]
&\leq \left(1+\frac{h}{d'}\right) \left(1+\frac{1}{(d')^{1/2-\eta}-1}\right) M_1\\
&+2 (d'+h) \exp\left(-\frac{(d')^{2\eta}}{8}\right) M_2,
\end{align*}
\end{lem}
\stepcounter{pf}
{\it Remark}: This lemma implies that the left side of (\ref{upp_kappa}) has an upper bound
\begin{eqnarray*}
\inf_{\kappa\in (0,1)}\sup_{d'\geq1}&&\!\!\!\!\!\!\!\!\!\left(\frac{1}{\kappa} \left(1+\frac{h}{d'}\right) M_1\right.\\
&&\!\!\!\!\left.+2(d'+h)\exp\left(-\frac{(1-\kappa)^2d'}{8}\right) M_2\right)
\end{eqnarray*}
that does not depend on $d$.

\begin{lem}
\label{lem_upp2}
Let $Z=(Z_1,\ldots,Z_d)$ denote a $d$-dimensional standard normal random vector $(d>4)$.
Let $\alpha,\beta,\gamma,\omega$ be natural numbers satisfying $1\leq\alpha\leq \beta\leq \gamma\leq\omega\leq d$.
Further, define $d'=d-4$. Then for arbitrary $\kappa \in(0,1)$, we have 
\begin{align}
E \Biggl[ & \left(\frac{d Z_\alpha Z_\beta Z_\gamma Z_\omega}{\|Z\|^2}\right)^2 \Biggr] \nonumber \\
\leq & \frac{1}{\kappa^2} \left(1+\frac{4}{d'}\right)^2 M_3  \nonumber \\
& +  2(d'+4)^2  \exp\left(-\frac{(1-\kappa)^2d'}{8}\right) M_4,
\label{upp_kappa2}
\end{align}
where we have defined $M_3$ and $M_4$ as
\begin{align*}
M_3 &=E\left[ (Z_\alpha Z_\beta Z_\gamma Z_\omega)^2  \right] \\
M_4 &=E\left[ (Z_\gamma Z_\omega)^2  \right].
\end{align*}
\end{lem}
{\it Remark}: This lemma implies that the left side of (\ref{upp_kappa2}) has an upper bound
\begin{align*}
\inf_{\kappa\in (0,1)}\sup_{d'\geq 1}
\Biggl( & \frac{1}{\kappa^2} \left(1+\frac{4}{d'}\right)^2 M_3  \\
 & 
  +2(d'+4)^2 \exp\left(-\frac{(1-\kappa)^2d'}{8}\right) M_4 \Biggr)
\end{align*}
that does not depend on $d$.

\begin{pf}
Define the random vector $Z'=(Z_{h+1},\ldots,Z_d)$
according to the $d'$-dimensional standard normal distribution, 
and the event $B = \{ \|Z'\|^2\leq \kappa d' \}$ for $\kappa\in (0,1)$.
Let $E_{B}$ denote the conditional expectation under $B$.
Then, we have 
\begin{eqnarray}
\!\!E\left[\frac{d\prod_{k=1}^h Z_k^{2m_k}}{\|Z\|^2} \right]
&&\!\!\!\!\!\!\!\!\!\!\!\!=\Pr(\bar{B})E_{\bar{B}}\!\!\left[\frac{d\prod_{k=1}^h Z_k^{2m_k}}{\|Z\|^2} \right]\nonumber\\
\!\!\!\!\!\!&&\!\!\!\!\!\!\!\!\!\!\!\!\!\!+\Pr(B)E_{B}\!\!\left[\frac{d\prod_{k=1}^h Z_k^{2m_k}}{\|Z\|^2} \right].
\label{upp-01}
\end{eqnarray}
In Lemma \ref{lem_prob}, substituting $(1-\kappa)\sqrt{d'}$ to $t$, we have
\begin{eqnarray*}
\Pr\left(\left|\frac{\|Z'\|^2}{d'}-1\right|\geq 1-\kappa\right)
\leq 2\exp\left(-\frac{(1-\kappa)^2{d'}}{8}\right).
\end{eqnarray*}
Thus, we have
\begin{eqnarray}
\Pr(B)&=&\Pr\left(\frac{\|Z'\|^2}{d'}\leq \kappa\right)\nonumber\\
&=&\Pr\left(\frac{\|Z'\|^2}{d'}-1\leq -(1-\kappa)\right)\nonumber\\
&\leq& \Pr\left(\left|\frac{\|Z'\|^2}{d'}-1\right|\geq 1-\kappa\right)\nonumber\\
&\leq& 2\exp\left(-\frac{(1-\kappa)^2{d'}}{8}\right).\label{upp-02}
\end{eqnarray}

Since 
$\bar{B}$ implies
$\|Z'\|^2\geq \kappa d'$ 
and since
$\|Z\|^2\geq \|Z'\|^2\geq \kappa d'$,
we have
\begin{eqnarray*}
\!\!\!\!\!\!\!\!E_{\bar{B}}\!\!\left[\frac{d\prod_{k=1}^h Z_k^{2m_k}}{\|Z\|^2} \right]
\!\!\!\!&\leq&\!\!\!\! E_{\bar{B}}\!\!\left[\frac{d\prod_{k=1}^h Z_k^{2m_k}}{\kappa d'} \right]\nonumber\\
\!\!\!\!&=&\!\!\!\!\frac{1}{\kappa}\left(1+\frac{h}{d'}\right)\prod_{k=1}^h E_{\bar{B}}\left[ Z_k^{2m_k } \right].
\end{eqnarray*}
Note that
\begin{eqnarray*}
\prod_{k=1}^h E_{\bar{B}}\left[ Z_k^{2m_k } \right]=\prod_{k=1}^h E\left[ Z_k^{2m_k } \right]=M_1
\end{eqnarray*}
holds,
since $\bar{B}$ is independent of $(Z_1,\dots,Z_h)$.
Thus, we have 
\begin{eqnarray}
E_{\bar{B}}\!\!\left[\frac{d\prod_{k=1}^h Z_k^{2m_k}}{\|Z\|^2} \right]
\!\!\leq\frac{1}{\kappa}\left(1+\frac{h}{d'}\right) M_1.\label{upp-03}
\end{eqnarray}

Next, consider the conditional expectation under $B$.
Using the fact $\|Z\|^2\geq Z_1^2$, we have 
\begin{eqnarray}
\!\!\!\!\!\!\!\!E_{B}\!\!\left[\frac{d\prod_{k=1}^h Z_k^{2m_k}}{\|Z\|^2} \right]
\!\!\!\!&\leq&\!\!\!\! E_{B}\!\!\left[\frac{d\prod_{k=1}^h Z_k^{2m_k}}{Z_1^2} \right]\nonumber\\
\!\!\!\!&=&\!\!\!\! dE\left[ Z_1^{2(m_1-1) } \right] \prod_{k=2}^h E\left[ Z_k^{2m_k } \right]\nonumber\\
\!\!\!\!&=&\!\!\!\! dM_2.\label{upp-04}
\end{eqnarray}
From (\ref{upp-01}), (\ref{upp-02}), (\ref{upp-03}), (\ref{upp-04}), and $\Pr(\bar{B})\leq 1$, we have 
\begin{eqnarray*}
&&\!\!\!\!\!\!\!\!\!\!\!\!E\left[\frac{d\prod_{k=1}^h Z_k^{2m_k}}{\|Z\|^2} \right]\\
&\leq&\!\!\!\!\frac{1}{\kappa} \left(1+\frac{h}{d'}\right) M_1
+2\exp\left(-\frac{(1-\kappa)^2d'}{8}\right)(d'+h) M_2.
\end{eqnarray*}

Here, let $1-\kappa=1/(d')^{1/2-\eta}$ with $\eta\in(0,1/2)$. Noting that
\begin{eqnarray*}
\frac{1}{\kappa}=\frac{1}{1-1/(d')^{1/2-\eta}}
=\frac{(d')^{1/2-\eta}}{(d')^{1/2-\eta}-1}
=1+\frac{1}{(d')^{1/2-\eta}-1},
\end{eqnarray*}
we have 
\begin{eqnarray*}
\!\!E\left[\frac{d\prod_{k=1}^h Z_k^{2m_k}}{\|Z\|^2} \right]
\!\!\!\!&\leq&\!\!\!\!\left(1+\frac{1}{(d')^{1/2-\eta}-1}\right) \left(1+\frac{h}{d'}\right) M_1\nonumber\\
\!\!\!\!&+&\!\!\!\!2\exp\left(-\frac{1}{(d')^{1-2\eta}}\cdot\frac{d'}{8}\right)(d'+h) M_2,
\end{eqnarray*}
which yields the claim of the lemma.
\end{pf}

\begin{pf}
Assume that $1\leq\alpha\leq \beta\leq \gamma\leq\omega\leq 4$ without loss of generality.
Define the $d'$-dimensional random vector $Z'=(Z_{5},\ldots,Z_d)$
and the event $B = \{\|Z'\|^2\leq \kappa d' \}$ for $\kappa\in (0,1)$.
Let $E_{B}$ denote the conditional expectation under $B$.
Then, we have 
\begin{eqnarray}
\!\!E\left[\left(\frac{dZ_\alpha Z_\beta Z_\gamma Z_\omega}{\|Z\|^2}\right)^2 \right]
&&\!\!\!\!\!\!\!\!\!\!\!\!=\Pr(\bar{B})E_{\bar{B}}
\!\!\left[\left(\frac{dZ_\alpha Z_\beta Z_\gamma Z_\omega}{\|Z\|^2}\right)^2 \right]\nonumber\\
\!\!\!\!\!\!&&\!\!\!\!\!\!\!\!\!\!\!\!\!\!\!\!\!\!\!
+\Pr(B)E_{B}\!\!\left[\left(\frac{dZ_\alpha Z_\beta Z_\gamma Z_\omega}{\|Z\|^2}\right)^2 \right].
\label{upp2-01}
\end{eqnarray}

Consider the conditional expectation under $\bar{B}$.
Since
$\|Z\|^2\geq \|Z'\|^2\geq \kappa d'$ holds under $\bar{B}$, we have
\begin{eqnarray}
&&\!\!\!\!\!\!\!\!\!\!\!\!E_{\bar{B}}\!\!\left[\left(\frac{dZ_\alpha Z_\beta Z_\gamma Z_\omega}{\|Z\|^2}\right)^2 \right]\nonumber\\
\!\!\!\!&\leq&\!\!\!\! E_{\bar{B}}\!\!\left[\left(\frac{dZ_\alpha Z_\beta Z_\gamma Z_\omega}{\kappa d'}\right)^2 \right]\nonumber\\
\!\!\!\!&=&\!\!\!\!\frac{1}{\kappa^2}\left(1+\frac{4}{d'}\right)^2  E_{\bar{B}}\left[ (Z_\alpha Z_\beta Z_\gamma Z_\omega)^2 \right]\nonumber\\
\!\!\!\!&=&\!\!\!\!\frac{1}{\kappa^2}\left(1+\frac{4}{d'}\right)^2  M_3.\label{upp2-03}
\end{eqnarray}

Next, consider the conditional expectation under $B$.
Using the fact 
\begin{eqnarray*}
\|Z\|^4=\left(\sum_{\alpha=1}^d Z_\alpha^2\right)^2 \geq Z_\alpha^2Z_\beta^2, 
\end{eqnarray*}
we have 
\begin{eqnarray}
\!\!\!\!\!\!\!\!\!\!\!\!E_{B}\!\!\left[\left(\frac{dZ_\alpha Z_\beta Z_\gamma Z_\omega}{\|Z\|^2}\right)^2 \right]
&\leq& E_{B}\!\!\left[\frac{\left(dZ_\alpha Z_\beta Z_\gamma Z_\omega\right)^2}{Z_\alpha^2Z_\beta^2} \right]\nonumber\\
\!\!\!\!&=&\!\!\!\! d^2   E\left[ \left( Z_\gamma Z_\omega\right)^2 \right]\nonumber\\
\!\!\!\!&=&\!\!\!\! d^2 M_4.\label{upp2-04}
\end{eqnarray}
From (\ref{upp-02}), (\ref{upp2-01}), (\ref{upp2-03}), (\ref{upp2-04}), and $\Pr(\bar{B})\leq 1$, we have 
\begin{eqnarray*}
\!\!\!\!&&\!\!\!\!\!\!\!\!\!\!\!\!E\left[\left(\frac{dZ_\alpha Z_\beta Z_\gamma Z_\omega}{\|Z\|^2}\right)^2 \right]\\
\!\!\!\!&\leq&\!\!\!\!\frac{1}{\kappa^2} \left(1+\frac{4}{d'}\right)^2 M_3
+2\exp\left(-\frac{(1-\kappa)^2d'}{8}\right)(d'+4)^2 M_4,
\end{eqnarray*}
which yields the lemma.
\end{pf}

\section{Lower bound on the expectation of some quantities}
\label{exp_low}
Next, we will give lower bounds on the expectation of 
some random variables
.
We have the following lemma.
\begin{lem}
\label{lem_low}
Let $Z=(Z_1,\ldots,Z_d)$ denote a $d$-dimensional standard normal random vector,
and define $d_1=d-2$ and $d_2=d-1$.
Then for arbitrary $\eta \in(0,1/2)$, we have 
\begin{eqnarray*}
E\left[\frac{d Z_1^2 Z_2^2}{\|Z\|^2} \right]\geq 1-\xi_1(d_1)
\end{eqnarray*}
and 
\begin{eqnarray*}
E\left[\frac{d Z_1^4}{\|Z\|^2} \right]\geq 3-2\xi_2(d_2),
\end{eqnarray*}
where the functions $\xi_1$ and $\xi_2$ are defined in Section \ref{eva_eigen}. 
\end{lem}

\begin{pf}
Let $E_Z$ denote the expectation for the random variable $Z$.
Define the random vector $Z'=(Z_2,\ldots,Z_d)$. Noting that $Z_1$ and $Z'$ are independent, we have
\begin{eqnarray*}
E_Z\left[\frac{ Z_1^2 Z_2^2}{\|Z\|^2}\right]
&=&E_{Z'} E_{Z_1}\left[\frac{Z_1^2 Z_2^2}{\|Z\|^2}\right]\\
&=&E_{Z'} \left[Z_2^2 E_{Z_1}\left[\frac{Z_1^2 }{Z_1^2+\|Z'\|^2}\right]\right].
\end{eqnarray*}
First, we evaluate $E_{Z} [Z_1^2/(Z_1^2+\|z'\|^2)]$ as 
\begin{eqnarray}
\!\!\!\!\!\!\!\!\!\!\!E_{Z_1} \left[\frac{Z_1^2}{Z_1^2+\|z'\|^2}\right]
&\geq&\int_{-a}^a\frac{z_1^2 p(z_1)}{z_1^2+\|z'\|^2}dz_1\nonumber\\
&\geq&\int_{-a}^a\frac{z_1^2 p(z_1)}{a^2+\|z'\|^2}dz_1\label{lemlow1}
\end{eqnarray}
with $a>0$.
Note that we have
\begin{eqnarray}
\!\!\int_{-a}^az_1^2 p(z_1)dz_1\!\!\!\!&=&\!\!\!\!1-2\int_{a}^\infty z_1^2 p(z_1)dz_1\nonumber\\
&=&\!\!\!\!1-\frac{2}{\sqrt{2\pi}}\int_{a}^\infty z_1^2 \exp\left(-\frac{z_1^2}{2}\right)dz_1\label{lemlow2}
\end{eqnarray}
and 
\begin{eqnarray}
&&\!\!\!\!\!\!\!\!\int_{a}^\infty z_1^2 \exp\left(-\frac{z_1^2}{2}\right) dz_1\nonumber\\
&=&\int_{a}^\infty z_1 \left(-\exp\left(-\frac{z_1^2}{2}\right)\right)' dz_1\nonumber\\
&=&\left[-z_1\exp\left(-\frac{z_1^2}{2}\right)\right]_a^\infty 
+\int_{a}^\infty \exp\left(-\frac{z_1^2}{2}\right) dz_1\nonumber\\
&\leq&a\exp\left(-\frac{a^2}{2}\right)+\int_{a}^\infty \frac{z_1}{a}\exp\left(-\frac{z_1^2}{2}\right) dz_1\nonumber\\
&=&a\exp\left(-\frac{a^2}{2}\right)+\left[-\frac{1}{a}\exp\left(-\frac{z_1^2}{2}\right)\right]_a^\infty\nonumber\\
&=&\left(a+\frac{1}{a}\right)\exp\left(-\frac{a^2}{2}\right)\label{lemlow3}
\end{eqnarray}
by integration by parts.
From (\ref{lemlow1}), (\ref{lemlow2}), and (\ref{lemlow3}), we have 
\begin{eqnarray*}
\!\!\!\!\!\!\!\!\!\!\!E_{Z_1} \left[\frac{Z_1^2}{Z_1^2+\|z'\|^2}\right]
\!\!&\geq&\!\!\frac{1-\iota_1(a)}{a^2+\|z'\|^2}.
\end{eqnarray*}
Thus, we have
\begin{eqnarray*}
E_Z\left[\frac{ Z_{1}^2 Z_{2}^2}{\|Z\|^2}\right]&\geq& 
(1-\iota_1(a))E_{Z'}\left[\frac{Z_2^2}{a^2+\|Z'\|^2}\right]\\
&=&(1-\iota_1(a))E_{Z''}E_{Z_2}\left[\frac{Z_2^2}{a^2+Z_2^2+\|Z''\|^2}\right],
\end{eqnarray*}
where we have defined the random vector $Z''=(Z_3,\ldots,Z_d)$.
By having the similar discussion on the expectation for $Z_2$, we have
\begin{eqnarray*}
E_{Z_2}\left[\frac{Z_2^2}{a^2+Z_2^2+\|z''\|^2}\right]&\geq& 
\int_{-a}^a\frac{z_2^2 p(z_2)}{a^2+z_2^2+\|z''\|^2}dz_2\\
&\geq& \int_{-a}^a\frac{z_2^2 p(z_2)}{2a^2+\|z''\|^2}dz_2\\
&\geq& \frac{1-\iota_1(a)}{2a^2+\|z''\|^2}.
\end{eqnarray*}
Thus, we have
\begin{eqnarray}
E_Z\left[\frac{ Z_{1}^2 Z_{2}^2}{\|Z\|^2}\right]\geq 
(1-\iota_1(a))^2E_{Z''}\left[\frac{1}{2a^2+\|Z''\|^2}\right]\label{eps}.
\end{eqnarray}
Letting $d_1=d-2$, we have from Lemma \ref{lem_prob}
\begin{eqnarray*}
\Pr\left(\left|\frac{\|Z''\|^2}{d_1}-1\right|\geq \frac{t}{\sqrt{d_1}}\right)\leq 2\exp\left(-\frac{t^2}{8}\right)
\end{eqnarray*}
with $t\in(0,\sqrt{d_1})$. Since 
\begin{eqnarray*}
&&\!\!\!\!\!\!\!\!\Pr\left(\left|\frac{\|Z''\|^2}{d_1}-1\right|\geq \frac{t}{\sqrt{d_1}}\right)\\
&\geq&\Pr\left(\frac{\|Z''\|^2}{d_1}-1\geq \frac{t}{\sqrt{d_1}}\right)\\
&=&\Pr\left(\|Z''\|^2\geq d_1+\sqrt{d_1}t\right)\\
&=&\Pr\left(\frac{1}{2a^2+\|Z''\|^2}\leq \frac{1}{d_1+\sqrt{d_1}t+2a^2}\right),
\end{eqnarray*}
we have
\begin{eqnarray*}
\Pr\left(\frac{1}{2a^2+\|Z''\|^2}\leq  \frac{1}{d_1+t\sqrt{d_1}+2a^2}\right)\leq 2\exp\!\left(-\frac{t^2}{8}\right).
\end{eqnarray*}
Considering the probability of the complementary event, we have
\begin{eqnarray*}
\Pr\!\left(\frac{1}{2a^2+\|Z''\|^2}\geq  \frac{1}{d_1+t\sqrt{d_1}+2a^2}\right)\geq 
1\!-\!2\exp\!\left(-\frac{t^2}{8}\right).
\end{eqnarray*}
Here, according to Markov's inequality, the probability of the left side is bounded by
\begin{eqnarray*}
E_{Z''}\left[\frac{1}{2a^2+\|Z''\|^2}\right](d_1+t\sqrt{d_1}+2a^2).
\end{eqnarray*}
Thus, we have
\begin{eqnarray}
E_{Z''}\left[\frac{1}{2a^2+\|Z''\|^2}\right]\geq \frac{1\!-\!2\exp\!\left(-\frac{t^2}{8}\right)}{d_1+t\sqrt{d_1}+2a^2}.\label{eps-a}
\end{eqnarray}
Applying the above inequality to (\ref{eps}), we have
\begin{eqnarray*}
E_Z\left[\frac{d Z_{1}^2 Z_{2}^2}{\|Z\|^2}\right]\geq(1-\iota_1(a))^2 
\frac{\left(1\!-\!2\exp\!\left(-\frac{t^2}{8}\right)\right)(d_1+2)}{d_1+t\sqrt{d_1}+2a^2}.
\end{eqnarray*}
Assuming that $a=d_1^\eta$ and $t=d_1^\eta$ for $\eta\in(0,1/2)$, 
we have  
\begin{eqnarray}
E_Z\left[\frac{d Z_{1}^2 Z_{2}^2}{\|Z\|^2}\right]\geq 1-\xi_1(d_1).
\end{eqnarray}

Next, we evaluate $E_{Z} [Z_1^4/\|Z\|^2]$ as 
\begin{eqnarray*}
E_Z\left[\frac{ Z_1^4}{\|Z\|^2}\right]
&=&E_{Z'} E_{Z_1}\left[\frac{Z_1^4}{Z_1^2 +\|Z'\|^2}\right].
\end{eqnarray*}
We evaluate $E_{Z} [Z_1^4/(Z_1^2+\|z'\|^2)]$ as 
\begin{eqnarray}
\!\!\!\!\!\!\!\!\!\!\!E_{Z_1} \left[\frac{Z_1^4}{Z_1^2+\|z'\|^2}\right]
&\geq&\int_{-a}^a\frac{z_1^4 p(z_1)}{z_1^2+\|z'\|^2}dz_1\nonumber\\
&\geq&\int_{-a}^a\frac{z_1^4 p(z_1)}{a^2+\|z'\|^2}dz_1\label{lemlow2-1}
\end{eqnarray}
with $a>0$.
Note that we have
\begin{eqnarray}
\!\!\int_{-a}^az_1^4 p(z_1)dz_1\!\!\!\!&=&\!\!\!\!3-2\int_{a}^\infty z_1^4 p(z_1)dz_1\nonumber\\
&=&\!\!\!\!3-\frac{2}{\sqrt{2\pi}}\int_{a}^\infty z_1^4 \exp\left(-\frac{z_1^2}{2}\right)dz_1\label{lemlow2-2}
\end{eqnarray}
and 
\begin{eqnarray}
&&\!\!\!\!\!\!\!\!\int_{a}^\infty z_1^4 \exp\left(-\frac{z_1^2}{2}\right) dz_1\nonumber\\
&=&\int_{a}^\infty z_1^3 \left(-\exp\left(-\frac{z_1^2}{2}\right)\right)' dz_1\nonumber\\
&=&\left[-z_1^3\exp\left(-\frac{z_1^2}{2}\right)\right]_a^\infty 
+\int_{a}^\infty 3z_1^2\exp\left(-\frac{z_1^2}{2}\right) dz_1\nonumber\\
&\leq&a^3\exp\left(-\frac{a^2}{2}\right)+3\left(a+\frac{1}{a}\right)\exp\left(-\frac{a^2}{2}\right)\nonumber\\
&\leq&\left(a^3+3a+\frac{3}{a}\right)\exp\left(-\frac{a^2}{2}\right)\label{lemlow2-3}
\end{eqnarray}
by integration by parts and (\ref{lemlow3}).
From (\ref{lemlow2-1}), (\ref{lemlow2-2}), and (\ref{lemlow2-3}), we have 
\begin{eqnarray*}
\!\!\!\!\!\!\!\!\!\!\!E_{Z_1} \left[\frac{Z_1^4}{Z_1^2+\|z'\|^2}\right]
\!\!&\geq&\!\!\frac{3-\iota_2(a)}{a^2+\|z'\|^2}.
\end{eqnarray*}
Thus, we have
\begin{eqnarray}
E_Z\left[\frac{ Z_{1}^4}{\|Z\|^2}\right]\geq 
(3-\iota_2(a))E_{Z'}\left[\frac{1}{a^2+\|Z'\|^2}\right]\label{eps2}.
\end{eqnarray}
By the same discussion  as for (\ref{eps-a}), we have
\begin{eqnarray*}
E_{Z'}\left[\frac{1}{a^2+\|Z'\|^2}\right]\geq \frac{1\!-\!2\exp\!\left(-\frac{t^2}{8}\right)}{d_2+t\sqrt{d_2}+a^2},
\end{eqnarray*}
where we have defined $d_2=d-1$.
Applying the above inequality to (\ref{eps2}), we have
\begin{eqnarray*}
E_Z\left[\frac{d Z_{1}^4}{\|Z\|^2}\right]\geq(3-\iota_2(a))
\frac{\left(1\!-\!2\exp\!\left(-\frac{t^2}{8}\right)\right)(d_2+1)}{d_2+t\sqrt{d_2}+a^2}.
\end{eqnarray*}
Assuming that $a=d_2^\eta$ and $t=d_2^\eta$ for $\eta\in(0,1/2)$, 
we have 
\begin{eqnarray*}
E_Z\left[\frac{d Z_{1}^4}{\|Z\|^2}\right]
\geq 3-2\xi_2(d_2).
\end{eqnarray*}
which yields the lemma.
\end{pf}

\section{Proof of Lemma \ref{lem_exp_var}}
\label{orth}
%
In this proof, let $E$ and $Var$ denote the expectation and the variance for the random variable $W$, 
respectively.
Further, we define the random vector
\begin{eqnarray*}
Z=\sqrt{p}(W_{1i},W_{2i}\ldots,W_{di}),
\end{eqnarray*}
and let $Z_\alpha$ denote the $\alpha$-th component of $Z$.
Note that $Z$ follows the $d$-dimensional standard normal distribution.
Now, we will prove 1), 2), and 3) in the lemma.
\begin{enumerate}
\item 
We can easily see that 
\begin{eqnarray*}
E[\|v^{(0)}\|^2]&=&\frac{1}{d}\sum_{i=1}^p E\left[\|W^{(i)}\|^2 \right]\\
&=&\frac{1}{d}\sum_{i=1}^p \sum_{l=1}^d E\left[W_{li}^2 \right]=1
\end{eqnarray*}
and
\begin{eqnarray*}
E[\|W_{l}\|^2]&=&\sum_{i=1}^p E[W_{li}^2]=p\cdot\frac{1}{p}=1.
\end{eqnarray*}
Calculating $E[\|v^{(\alpha,\beta)}\|^2]$, we have
\begin{eqnarray}
E[\|v^{(\alpha,\beta)}\|^2]&=&\sum_{i=1}^p E\left[\frac{d W_{\alpha i}^2 W_{\beta i}^2}{\|W^{(i)}\|^2}\right]\nonumber\\
&=&E\left[\frac{d Z_{\alpha}^2 Z_{\beta}^2}{\|Z\|^2}\right].\label{e_alpha_beta}
\end{eqnarray}
Since it is difficult to explicitly calculate the above expectation, we
will give upper and lower bounds on it.
Assume that $\alpha=1$ and $\beta=2$ without loss of generality.
Letting $h=2$, $m_1=1$, and $m_2=1$ in Lemma \ref{lem_upp}, we have for arbitrary $\eta \in(0,1/2)$
\begin{eqnarray}
\!\!\!\!\!\!\!\!E\left[\frac{d Z_1^{2}Z_2^{2}}{\|Z\|^2} \right]
&\leq&\left(1+\frac{2}{d_1}\right) \left(1+\frac{1}{(d_1)^{1/2-\eta}-1}\right)\nonumber\\
&&+2(d_1+2) \exp\left(-\frac{(d_1)^{2\eta}}{8}\right)\nonumber\\
&=& 1+\bar{\xi}_1(d_1),\label{exp_upp1}
\end{eqnarray}
where we have defined $d_1=d-2$.
Further from Lemma \ref{lem_low}, we have 
\begin{eqnarray}
E\left[\frac{d Z_1^2 Z_2^2}{\|Z\|^2} \right]\geq 1-\xi_1(d_1).\label{exp_low1}
\end{eqnarray}
From (\ref{exp_upp1}) and (\ref{exp_low1}), we have 
\begin{eqnarray*}
\left|E[\|v^{(\alpha,\beta)}\|^2]-1\right|\leq \xi(d).
\end{eqnarray*}

Next, we evaluate
\begin{eqnarray}
&&\!\!\!\!\!\!\!\!\!\!E[\|v^{(\gamma)}\|^2]\nonumber\\
&=&\frac{1}{2}E[\|v^{(\gamma,\gamma)}-v^{(0)}\|^2]\nonumber\\
&=&\frac{1}{2}E[\|v^{(\gamma,\gamma)}\|^2-2v^{(\gamma,\gamma)}\cdot v^{(0)} +\|v^{(0)}\|^2]\nonumber\\
&=&\frac{1}{2}\left(E[\|v^{(\gamma,\gamma)}\|^2]-1\right)\label{vgamma}
\end{eqnarray}
where we have used 
\begin{eqnarray*}
E[v^{(\gamma,\gamma)}\cdot v^{(0)}]&=&\sum_{i=1}^p E\left[\frac{\sqrt{d}W_{\gamma i}^2}{\|W^{(i)}\|}\cdot 
\frac{\|W^{(i)}\|}{\sqrt{d}} \right]\\
&=&\sum_{i=1}^p E[W_{\gamma i}^2]=1.
\end{eqnarray*}
Now we evaluate
\begin{eqnarray}
E[\|v^{(\gamma,\gamma)}\|^2]&=&\sum_{i=1}^p E\left[\frac{dW_{\gamma i}^4}{\|W^{(i)}\|^2}\right]\nonumber\\
&=&E\left[\frac{d Z_{\gamma}^4}{\|Z\|^2}\right].\nonumber
\end{eqnarray}
Assume that $\gamma=1$ without loss of generality.
Letting $h=1$, and $m_1=2$ in Lemma \ref{lem_upp}, we have for arbitrary $\eta \in(0,1/2)$
\begin{eqnarray}
\!\!\!\!\!\!\!\!E\left[\frac{d Z_1^{4}}{\|Z\|^2} \right]
&\leq&3\left(1+\frac{1}{d_2}\right)\left(1+\frac{1}{(d_2)^{1/2-\eta}-1}\right)\nonumber\\
&&+2(d_2+1) \exp\left(-\frac{(d_2)^{2\eta}}{8}\right)\nonumber\\
&=&3+2\bar{\xi}_2(d_2),\label{exp_upp2}
\end{eqnarray}
where we have defined $d_2=d-1$.
Further from Lemma \ref{lem_low}, we have 
\begin{eqnarray}
E\left[\frac{d Z_1^4}{\|Z\|^2} \right]\geq 3-2\xi_2(d_2).\label{exp_low2}
\end{eqnarray}
From (\ref{exp_upp2}) and (\ref{exp_low2}), we have 
\begin{eqnarray*}
\left|E[\|v^{(\gamma,\gamma)}\|^2]-3\right|\leq 2\xi(d).
\end{eqnarray*}
From (\ref{vgamma}), we have 
\begin{eqnarray}
&&\left|2E[\|v^{(\gamma)}\|^2]+1-3\right|\leq 2\xi(d)\nonumber\\
&\Leftrightarrow&\left|E[\|v^{(\gamma)}\|^2]-1\right|\leq \xi(d)\label{vgammauplow}.
\end{eqnarray}

\item 
First, we examine the inner product of $v^{(0)}$ and the other vectors.
\begin{eqnarray*}
E[v^{(0)}\cdot W_l]&=&\sum_{i=1}^p E\left[\frac{1}{\sqrt{d}}\|W^{(i)}\| \cdot W_{li}\right]\\
&=&\frac{1}{\sqrt{d}}E\left[\|Z\|Z_{l}\right]\\
&=&\frac{1}{\sqrt{d}}\int_{\Re^d}\|z\|z_l\ p(z)dz=0.
\end{eqnarray*}
The above holds because the integrand is an odd function with respect to $z_l$, 
thus the integral equals zero.
Further, we have
\begin{eqnarray*}
E[v^{(0)}\cdot v^{(\alpha,\beta)}]&=&\sum_{i=1}^p E\left[\frac{\|W^{(i)}\|}{\sqrt{d}}
\frac{\sqrt{d}W_{\alpha i}W_{\beta i}}{\|W^{(i)}\|}\right]\\
&=&\sum_{i=1}^pE\left[W_{\alpha i}W_{\beta i}\right]=0,
\end{eqnarray*}
and
\begin{eqnarray*}
E[v^{(0)}\cdot v^{(\gamma)}]
&=&\frac{1}{\sqrt{2}}E\left[v^{(0)}\cdot v^{(\gamma,\gamma)}-\|v^{(0)}\|^2\right]\\
&=&\frac{1}{\sqrt{2}}(1-1)=0.
\end{eqnarray*}

Secondly, we examine the inner product of $W_l$ and the other vectors:
\begin{eqnarray*}
E[W_l\cdot W_{l'}]&=&\sum_{i=1}^p E[W_{li}W_{l'i}]=0\ (l\neq l'),
\end{eqnarray*}
\begin{eqnarray*}
E[W_l\cdot v^{(\alpha,\beta)}]&=&\sum_{i=1}^p E\left[W_{li}\frac{\sqrt{d}W_{\alpha i}W_{\beta i}}{\|W^{(i)}\|}\right]\\
&=&\sqrt{d}E\left[\frac{Z_{l}Z_{\alpha}Z_{\beta}}{\|Z\|}\right]\\
&=&\sqrt{d}\int_{\Re^d}\frac{z_lz_{\alpha}z_{\beta}}{\|z\|}p(z)dz=0.
\end{eqnarray*}
The above holds because the integrand is an odd function with respect to $z_l$, $z_\alpha$, or $z_\beta$,
thus the integral equals zero. 
Similarly, we can see that
\begin{eqnarray*}
E[W_l\cdot v^{(\gamma,\gamma)}]=0.
\end{eqnarray*}
Thus, we have
\begin{eqnarray*}
E[W_l\cdot v^{(\gamma)}]
&=&\frac{1}{\sqrt{2}}E[W_l\cdot v^{(\gamma,\gamma)}-W_l\cdot v^{(0)}]\\
&=&\frac{1}{\sqrt{2}}(0-0)=0.
\end{eqnarray*}

Thirdly, we examine the inner product of $v^{(\alpha,\beta)}$ and the other vectors.
For the vector $v^{(\alpha',\beta')}$ where $\alpha'\neq \alpha$ or $\beta'\neq \beta$,
we have 
\begin{eqnarray*}
E[v^{(\alpha,\beta)}\cdot v^{(\alpha',\beta')}]\!\!\!\!&=&\!\!\!\!
\sum_{i=1}^p E\left[\frac{dW_{\alpha i}W_{\beta i}W_{\alpha' i}W_{\beta' i}}{\|W^{(i)}\|^2}\right]\\
&=&\!\!dE\left[\frac{Z_{\alpha}Z_{\beta}Z_{\alpha'}Z_{\beta'}}{\|Z\|^2}\right]\\
&=&\!\!d\int_{\Re^d}\frac{z_{\alpha}z_{\beta}z_{\alpha'}z_{\beta'}}{\|z\|^2}p(z)dz=0.
\end{eqnarray*}
Recalling that $\alpha<\beta$ and $\alpha'<\beta'$, the integrand is an odd function 
because all of the following cases are negated: (i) $\alpha=\beta$ and $\alpha'=\beta'$
 (ii) $\alpha=\alpha'$ and $\beta=\beta'$ (iii)  $\alpha=\beta'$ and $\alpha'=\beta$. 
Similarly, we can see that
\begin{eqnarray*}
E[v^{(\alpha,\beta)}\cdot v^{(\gamma,\gamma)}]=0.
\end{eqnarray*}
Thus, we have
\begin{eqnarray*}
&&\!\!\!\!\!\!\!\!E[v^{(\alpha,\beta)}\cdot v^{(\gamma)}]\\
&=&\frac{1}{\sqrt{2}}E\left[v^{(\alpha,\beta)}\cdot v^{(\gamma,\gamma)}-v^{(\alpha,\beta)}\cdot v^{(0)} \right]=0.
\end{eqnarray*}

Lastly, we examine the inner product of $v^{(\gamma)}$ and $v^{(\gamma')}$ ($\gamma'\neq \gamma$). 
From equation (\ref{sumgamma}), we have
\begin{eqnarray*}
&&\!\!\!\!\!\!\!\!E\left[\left\|\sum_{\gamma=1}^d v^{(\gamma)}\right\|^2\right]\\
&=&\!\!E\left[\sum_{\gamma=1}^d \|v^{(\gamma)}\|^2 + \sum_{\gamma\neq\gamma'}^d v^{(\gamma)}\cdot v^{(\gamma')} \right]\\
&=&\!\!dE\left[\|v^{(\gamma)}\|^2\right]+d(d-1)E\left[v^{(\gamma)}\cdot v^{(\gamma')} \right]=0.
\end{eqnarray*}
Thus, we have 
\begin{eqnarray*}
E\left[v^{(\gamma)}\cdot v^{(\gamma')}\right]=-\frac{1}{d-1}E\left[\|v^{(\gamma)}\|^2\right].
\end{eqnarray*}
Using (\ref{vgammauplow}), we have
\begin{eqnarray*}
\left|E[v^{(\gamma)}\cdot v^{(\gamma')}]-\left(-\frac{1}{d-1}\right)\right|\leq \frac{\xi(d)}{d-1}.
\end{eqnarray*}

\item First, we examine the variances of the norms of vectors.
We can easily see that 
\begin{eqnarray*}
Var\left[\|v^{(0)}\|^2\right]&=&\frac{1}{d^2}\sum_{i=1}^p Var\left[\|W^{(i)}\|^2 \right]\\
&=&\frac{1}{d^2}\sum_{i=1}^p \sum_{l=1}^d Var\left[W_{li}^2 \right]=\frac{2}{dp}.
\end{eqnarray*}
and
\begin{eqnarray*}
Var\left[\|W_{l}\|^2\right]&=&\sum_{i=1}^p Var\left[W_{li}^2\right]=p\cdot\frac{2}{p^2}=\frac{2}{p}.
\end{eqnarray*}

Similar to the calculation in (\ref{e_alpha_beta}), we have
\begin{eqnarray*}
Var\left[\|v^{(\alpha,\beta)}\|^2\right]&=&\sum_{i=1}^p Var\left[\frac{d W_{\alpha i}^2 W_{\beta i}^2}{\|W^{(i)}\|^2}\right]\\\
&=&\frac{1}{p}Var\left[\frac{d Z_{\alpha}^2 Z_{\beta}^2}{\|Z\|^2}\right]\\
&\leq&\frac{1}{p}E\left[\left(\frac{d Z_{\alpha}^2 Z_{\beta}^2}{\|Z\|^2}\right)^2\right].
\end{eqnarray*}
From Lemma \ref{lem_upp2}, the above expectation is bounded by the positive constant $C_1$.
Thus, we have
\begin{eqnarray*}
Var\left[\|v^{(\alpha,\beta)}\|^2\right]\leq \frac{C_1}{p}.
\end{eqnarray*}

Next, we evaluate 
\begin{eqnarray*}
&&\!\!\!\!\!\!\!\!\!\!\!\!Var\left[\|v^{(\gamma)}\|^2\right]\\
&=&\frac{1}{4}Var\left[\|v^{(\gamma,\gamma)}- v^{(0)}\|^2 \right]\\
\!\!\!\!&=&\!\!\!\!\frac{1}{4} \sum_{i=1}^p Var\left[\left(\frac{\sqrt{d}W_{\gamma i}^2}{\|W^{(i)}\|}-\frac{\|W^{(i)}\|}{\sqrt{d}}\right)^2\right]\\
\!\!\!\!&=&\!\!\!\!\frac{1}{4p} Var\left[\left(\frac{\sqrt{d}Z_{\gamma}^2}{\|Z\|}-\frac{\|Z\|}{\sqrt{d}}\right)^2\right]\\
\!\!\!\!&=&\!\!\!\!\frac{1}{4p} Var\left[\frac{dZ_{\gamma}^4}{\|Z\|^2}+\frac{\|Z\|^2}{d}-2Z_{\gamma}^2\right]\\
\!\!\!\!&\leq&\!\!\!\!\frac{1}{4p} E\left[\left(\frac{dZ_{\gamma}^4}{\|Z\|^2}+\frac{\|Z\|^2}{d}-2Z_{\gamma}^2
\right)^2\right]\\
\!\!\!\!&\leq&\!\!\!\!\frac{1}{4p} E\left[\left(\frac{dZ_{\gamma}^4}{\|Z\|^2}\right)^2\right]
\\&&\!\!\!\!+\frac{1}{4p} E\left[\frac{\|Z\|^4}{d^2}+\left(2Z_{\gamma}^2\right)^2 +2Z_{\gamma}^2\right]\\
\!\!\!\!&=&\!\!\!\!
\frac{1}{4p} E\left[\left(\frac{dZ_{\gamma}^4}{\|Z\|^2}\right)^2\right]+
\left(\frac{19}{4}+\frac{1}{2d}\right)\frac{1}{p},
\end{eqnarray*}
where we have used 
$E\left[\|Z\|^4\right]=d(d+2)$.
From Lemma \ref{lem_upp2}, 
the expectation 
$E\big[\left(dZ_{\gamma}^4/\|Z\|^2\right)^2\big]$
is bounded by the positive constant $C_2$.
Thus, we have
\begin{eqnarray*}
Var\left[\|v^{(\gamma)}\|^2\right]
\leq\left(\frac{C_2}{4}+\frac{19}{4}+\frac{1}{2d}\right)\frac{1}{p}.
\end{eqnarray*}

Now, we examine the variances of the inner products of the vectors.
First, we examine the inner product of $v^{(0)}$ and the other vectors:
\begin{eqnarray*}
Var\left[v^{(0)}\cdot W_l \right]\!\!\!\!&=&\!\!\sum_{i=1}^p Var\left[\frac{1}{\sqrt{d}}\|W^{(i)}\|\cdot W_{li}\right]\\
&=&\!\!\frac{1}{d}\sum_{i=1}^p  Var\left[ \|W^{(i)}\|W_{li}\right]\\
&=&\!\!\frac{1}{d}\sum_{i=1}^p  E\left[ \|W^{(i)}\|^2W_{li}^2\right]\\
&=&\!\!\frac{1}{d}\sum_{i=1}^p  E\left[W_{li}^4 +\sum_{k\neq l} W_{ki}^2 W_{l i}^2\right]\\
&=&\!\!\frac{p}{d}\left(\frac{3}{p^2}+ \frac{d-1}{p^2}\right)=\left(1+\frac{2}{d}\right)\frac{1}{p},
\end{eqnarray*}
\begin{eqnarray*}
Var\left[v^{(0)}\cdot v^{(\alpha,\beta)}\right]
\!\!\!\!&=&\!\!\sum_{i=1}^p Var\left[\frac{\|W^{(i)}\|}{\sqrt{d}}\frac{\sqrt{d}W_{\alpha i}W_{\beta i}}{\|W^{(i)}\|}\right]\\
&=&\!\!\sum_{i=1}^pVar\left[W_{\alpha i}W_{\beta i}\right]=\frac{1}{p},
\end{eqnarray*}
\begin{eqnarray*}
&&\!\!\!\!\!\!\!\!Var\left[v^{(0)}\cdot v^{(\gamma)}\right]\\
\!\!\!\!&=&\!\!\frac{1}{2}Var\left[v^{(0)}\cdot v^{(\gamma,\gamma)}-\|v^{(0)}\|^2\right]\\
&=&\!\!\frac{1}{2}\sum_{i=1}^p Var\left[W_{\gamma i}^2-\frac{1}{d}\|W^{(i)}\|^2\right]\\
&=&\!\!\frac{1}{2d^2}\sum_{i=1}^p Var\left[(d-1)W_{\gamma i}^2-\sum_{l\neq \gamma}W_{li}^2\right]\\
&=&\!\!\frac{p}{2d^2}\left(\frac{(d-1)^2}{p^2}+\frac{d-1}{p^2}\right)\\
&=&\!\!\frac{1}{2p}\left(1-\frac{1}{d}\right).
\end{eqnarray*}

Secondly, we examine the inner product of $W_i$ and the other vectors:
\begin{eqnarray*}
Var\left[W_l\cdot W_{l'}\right]&=&\sum_{i=1}^p Var\left[W_{li}W_{l'i}\right]=\frac{1}{p}\ (l\neq l'),
\end{eqnarray*}
\begin{eqnarray*}
Var\left[W_l\cdot v^{(\alpha,\beta)}\right]&=&\sum_{i=1}^p Var\left[W_{li}\frac{\sqrt{d}W_{\alpha i}W_{\beta i}}{\|W^{(i)}\|}\right]\\
&=&\frac{1}{p}E\left[\frac{dZ_{l}^2 Z_{\alpha}^2 Z_{\beta}^2}{\|Z\|^2}\right].
\end{eqnarray*}
From Lemma \ref{lem_upp}, the above expectation is bounded by the positive constant $C_3$.
Thus, we have
\begin{eqnarray*}
Var\left[W_l\cdot v^{(\alpha,\beta)}\right]\leq \frac{C_3}{p}.
\end{eqnarray*}
Next, we have
\begin{eqnarray*}
&&\!\!\!\!\!\!\!\!Var\left[W_l\cdot v^{(\gamma)}\right]\\
&=&\frac{1}{2}Var\left[W_l\cdot(v^{(\gamma,\gamma)}-v^{(0)})\right]\\
&=&\frac{1}{2}\sum_{i=1}^p Var\left[W_{li}\left(\frac{\sqrt{d}W_{\gamma i}^2}{\|W^{(i)}\|}-\frac{\|W^{(i)}\|}{\sqrt{d}}\right)\right]\\
&\leq&\frac{1}{2p} E\left[Z_{l}^2\left(\frac{\sqrt{d}Z_{\gamma}^2}{\|Z\|}-\frac{\|Z\|}{\sqrt{d}}\right)^2\right]\\
&=&\frac{1}{2p} E\left[Z_{l}^2\left(\frac{dZ_{\gamma}^4}{\|Z\|^2}+\frac{\|Z\|^2}{d}-2Z_{\gamma}^2\right)\right]\\
&=&\frac{1}{2p} E\left[\frac{dZ_{l}^2Z_{\gamma}^4}{\|Z\|^2}+\frac{Z_{l}^2\|Z\|^2}{d}-2Z_{l}^2Z_{\gamma}^2\right].
\end{eqnarray*}
From Lemma \ref{lem_upp}, the expectation $E\left[dZ_{l}^2 Z_{\gamma}^4/\|Z\|^2\right]$
is bounded by the positive constant $C_4$.
Thus, we have
\begin{eqnarray*}
Var\left[W_l\cdot v^{(\gamma)}\right]\leq \left(\frac{C_4}{2}-\frac{1}{2}+\frac{1}{d}\right)\frac{1}{p},
\end{eqnarray*}
where we have used 
$
E\left[Z_{l}^2\|Z\|^2\right]=d+2.
$


Thirdly, we examine the inner product of $v^{(\alpha,\beta)}$ and the other vectors:
\begin{eqnarray*}
&&\!\!\!\!\!\!\!\!Var\left[v^{(\alpha,\beta)}\cdot v^{(\alpha',\beta')}\right]\\
&=&\sum_{i=1}^p Var\left[\frac{dW_{\alpha i}W_{\beta i}W_{\alpha' i}W_{\beta' i}}{\|W^{(i)}\|^2}\right]\\
&=&\frac{1}{p} E\left[\left(\frac{dZ_{\alpha}Z_{\beta}Z_{\alpha'}Z_{\beta'}}{\|Z\|^2}\right)^2\right].
\end{eqnarray*}
From Lemma \ref{lem_upp2}, the above expectation is bounded by the positive constant $C_5$.
Thus, we have
\begin{eqnarray*}
Var\left[v^{(\alpha,\beta)}\cdot v^{(\alpha',\beta')}\right]
&\leq&\frac{C_5}{p}.
\end{eqnarray*}
Next, we have
\begin{eqnarray*}
&&\!\!\!\!\!\!\!\!\!\!\!\!Var\left[v^{(\alpha,\beta)}\cdot v^{(\gamma)}\right]\\
\!\!\!\!&=&\!\!\!\!\frac{1}{2}Var\left[v^{(\alpha,\beta)}\cdot (v^{(\gamma,\gamma)}- v^{(0)}) \right]\\
\!\!\!\!&=&\!\!\!\!\frac{1}{2}\sum_{i=1}^p Var\!\!\left[\frac{\sqrt{d}W_{\alpha i}W_{\beta i}}{\|W^{(i)}\|}\!
\left(\frac{\sqrt{d}W_{\gamma i}^2}{\|W^{(i)}\|}-\frac{\|W^{(i)}\|}{\sqrt{d}}\right)\right]\\
\!\!\!\!&\leq&\!\!\!\!\frac{1}{2p}E\!\!\left[\frac{dZ_{\alpha}^2Z_{\beta}^2}{\|Z\|^2}
\!\left(\frac{\sqrt{d}Z_{\gamma}^2}{\|Z\|}-\frac{\|Z\|}{\sqrt{d}}\right)^2\right]\\
\!\!\!\!&=&\!\!\!\!\frac{1}{2p}E\!\!\left[\left(\frac{d Z_{\alpha}Z_{\beta}Z_{\gamma}^2}{\|Z\|^2}\right)^2+Z_{\alpha}^2Z_{\beta}^2
-2\frac{dZ_{\alpha}^2Z_{\beta}^2Z_{\gamma}^2}{\|Z\|^2}\right]\\
\!\!\!\!&\leq&\!\!\!\!\frac{1}{2p}E\!\!\left[\left(\frac{d Z_{\alpha}Z_{\beta}Z_{\gamma}^2}{\|Z\|^2}\right)^2+Z_{\alpha}^2Z_{\beta}^2\right].
\end{eqnarray*}
From Lemma \ref{lem_upp2}, the expectation $E\left[\left(d Z_{\alpha}Z_{\beta}Z_{\gamma}^2/\|Z\|^2\right)^2\right]$
is bounded by the positive constant $C_6$.
Thus, we have
\begin{eqnarray*}
Var\left[v^{(\alpha,\beta)}\cdot v^{(\gamma)}\right]\leq \left(\frac{C_6+1}{2}\right)\frac{1}{p}.
\end{eqnarray*}

Lastly, we examine the inner product of $v^{(\gamma)}$ and $v^{(\gamma')}$ ($\gamma'\neq \gamma$). 
\begin{eqnarray*}
&&\!\!\!\!\!\!\!\!\!\!\!\!Var\left[v^{(\gamma)}\cdot v^{(\gamma')}\right]\\
&=&\frac{1}{4}Var\left[(v^{(\gamma,\gamma)}- v^{(0)})\cdot (v^{(\gamma',\gamma')}- v^{(0)}) \right]\\
\!\!\!\!&=&\!\!\!\!\frac{1}{4p} Var\left[\left(\frac{\sqrt{d}Z_{\gamma}^2}{\|Z\|}-\frac{\|Z\|}{\sqrt{d}}\right)\!
\left(\frac{\sqrt{d}Z_{\gamma'}^2}{\|Z\|}-\frac{\|Z\|}{\sqrt{d}}\right)\right]\\
\!\!\!\!&=&\!\!\!\!\frac{1}{4p} Var\left[\frac{dZ_{\gamma}^2Z_{\gamma'}^2}{\|Z\|^2}+\frac{\|Z\|^2}{d}-(Z_{\gamma}^2+Z_{\gamma'}^2)\right]\\
\!\!\!\!&\leq&\!\!\!\!\frac{1}{4p} E\left[\left(\frac{dZ_{\gamma}^2Z_{\gamma'}^2}{\|Z\|^2}+\frac{\|Z\|^2}{d}-(Z_{\gamma}^2+Z_{\gamma'}^2)
\right)^2\right]\\
\!\!\!\!&\leq&\!\!\!\!\frac{1}{4p} E\left[\left(\frac{dZ_{\gamma}^2Z_{\gamma'}^2}{\|Z\|^2}\right)^2\right]
\\&&\!\!\!\!+\frac{1}{4p} E\left[\frac{\|Z\|^4}{d^2}+\left(Z_{\gamma}^2+Z_{\gamma'}^2\right)^2 +2Z_{\gamma}^2Z_{\gamma'}^2\right]\\
\!\!\!\!&\leq&\!\!\!\!\left(\frac{C_1}{4}+\frac{11}{4}+\frac{1}{2d}\right)\frac{1}{p},
\end{eqnarray*}
where we have used 
$E\left[\|Z\|^4\right]=d(d+2)$.

By the discussion so far, we have confirmed that 
for all vectors $v,v'\in V$, the variance of the inner product is bounded as
\begin{eqnarray*}
Var\left[v\cdot v'\right]\leq \frac{C}{p},
\end{eqnarray*}
where $C$ is a certain positive constant.

\end{enumerate}


\ifCLASSOPTIONcaptionsoff
  \newpage
\fi

\end{document}